\documentclass[conference]{IEEEtran}
\usepackage{times}
\let\labelindent\relax

\usepackage[numbers]{natbib}
\usepackage{multicol}
\usepackage[bookmarks=true,colorlinks]{hyperref}
\usepackage{url}
\usepackage{algorithm}
\usepackage[noend]{algorithmic}
\usepackage[font={small}]{caption}
\hypersetup{colorlinks,allcolors=[rgb]{0,0.07843,0.45098}, urlcolor=blue}
\usepackage{blindtext}

\usepackage{bigstrut}
\usepackage{caption}
\usepackage{comment}
\usepackage{graphicx}
\usepackage{amsmath}

\def\1{\bm{1}}

\def\rd{{\textnormal{d}}}

\newcommand{\norm}[1]{\ensuremath{\left\| #1 \right\|}}

\usepackage{amsmath, amsfonts, bm, amssymb}
\usepackage{latexsym, mathtools}
\usepackage{graphicx}
\usepackage{float}
\usepackage{ifthen}
\usepackage{cleveref}
\usepackage{thmtools, thm-restate}
\usepackage{enumitem}

\usepackage{calc}
\usepackage{tikz}

\DeclareMathOperator{\bmtx}{\begin{bmatrix}}
\DeclareMathOperator{\emtx}{\end{bmatrix}}

\def\nd/{\textsuperscript{nd}}
\def\rd/{\textsuperscript{rd}}
\def\th/{\textsuperscript{th}}

\usepackage{titlesec}
\usepackage{booktabs}

\titlespacing\section{0pt}{5pt plus 1pt minus 1pt}{4pt plus 1pt minus 1pt}
\titlespacing\subsection{0pt}{4pt plus 1pt minus 1pt}{4pt plus 1pt minus 1pt}

\newcommand{\METHOD}{{\texttt{PoCo}}}

\begin{document}


\title{PoCo:  Policy Composition from and for Heterogeneous Robot Learning}

\author{Lirui Wang, Jialiang  Zhao, Yilun Du, Edward H. Adelson, Russ Tedrake \\ MIT CSAIL \\ \href{https://liruiw.github.io/policycomp}{\texttt{https://liruiw.github.io/policycomp}} }

\maketitle

\begin{abstract}
    Training general robotic policies from heterogeneous data for different tasks is a significant challenge. Existing robotic datasets vary in different modalities such as color, depth, tactile, and proprioceptive information, and collected in different domains such as simulation, real robots, and human videos. Current methods usually collect and pool all data from one domain to train a single policy to handle such heterogeneity in tasks and domains, which is prohibitively expensive and difficult. In this work, we present a flexible approach, dubbed Policy Composition, to combine information across such diverse modalities and domains for learning scene-level and task-level generalized manipulation skills, by composing different data distributions represented with diffusion models. Our method can use task-level composition for multi-task manipulation and be composed with analytic cost functions to adapt policy behaviors at inference time. We train our method on simulation, human, and real robot data and evaluate in tool-use tasks. The composed policy achieves robust and dexterous performance under varying scenes and tasks and outperforms baselines from a single data source and simple baselines that pool very heterogeneous data together in both simulation and real-world experiments. 

\end{abstract}

\IEEEpeerreviewmaketitle

\section{Introduction}
\label{sec:intro}


Training generalist robotic policies with massive amounts of computational power and data using simple algorithms such as behavior cloning \cite{osa2018algorithmic,bagnell2015invitation} has gathered considerable interest recently \cite{open_x_embodiment_rt_x_2023,jang2022bc}, inspired by the success of foundation models in fields such as natural language processing and computer vision \cite{kirillov2023segment,brown2020language,radford2021learning}. Despite the growing size of robotic datasets \cite{open_x_embodiment_rt_x_2023}, the dominant robot learning pipeline is still training specialist models with a single robot for a single task without any flexibility in behaviors \cite{wang2023mimicplay,chi2023diffusion,wang2022goal}. One of the key challenges for robot imitation learning under multi-task and multi-domain data at scale is data heterogeneity: in addition to the vast amount of tasks that robots can be trained to perform, robotic datasets were usually built from different data sources, such as simulations, human demonstrations, and real-robot data with different modalities such as color images, depth images, and tactile data. In this paper, we will refer to settings that accept these varieties of modalities, embodiments and tasks as \textbf{heterogenous robot learning}.
The efficient use of such data is not obvious due to such heterogeneity and most existing methods often only leverage one source of data such as real-world data \cite{open_x_embodiment_rt_x_2023}  or simulation data \cite{zhao2020sim,peng2018sim}.

\begin{figure}[!t]
\centering 
\includegraphics[width=\linewidth]{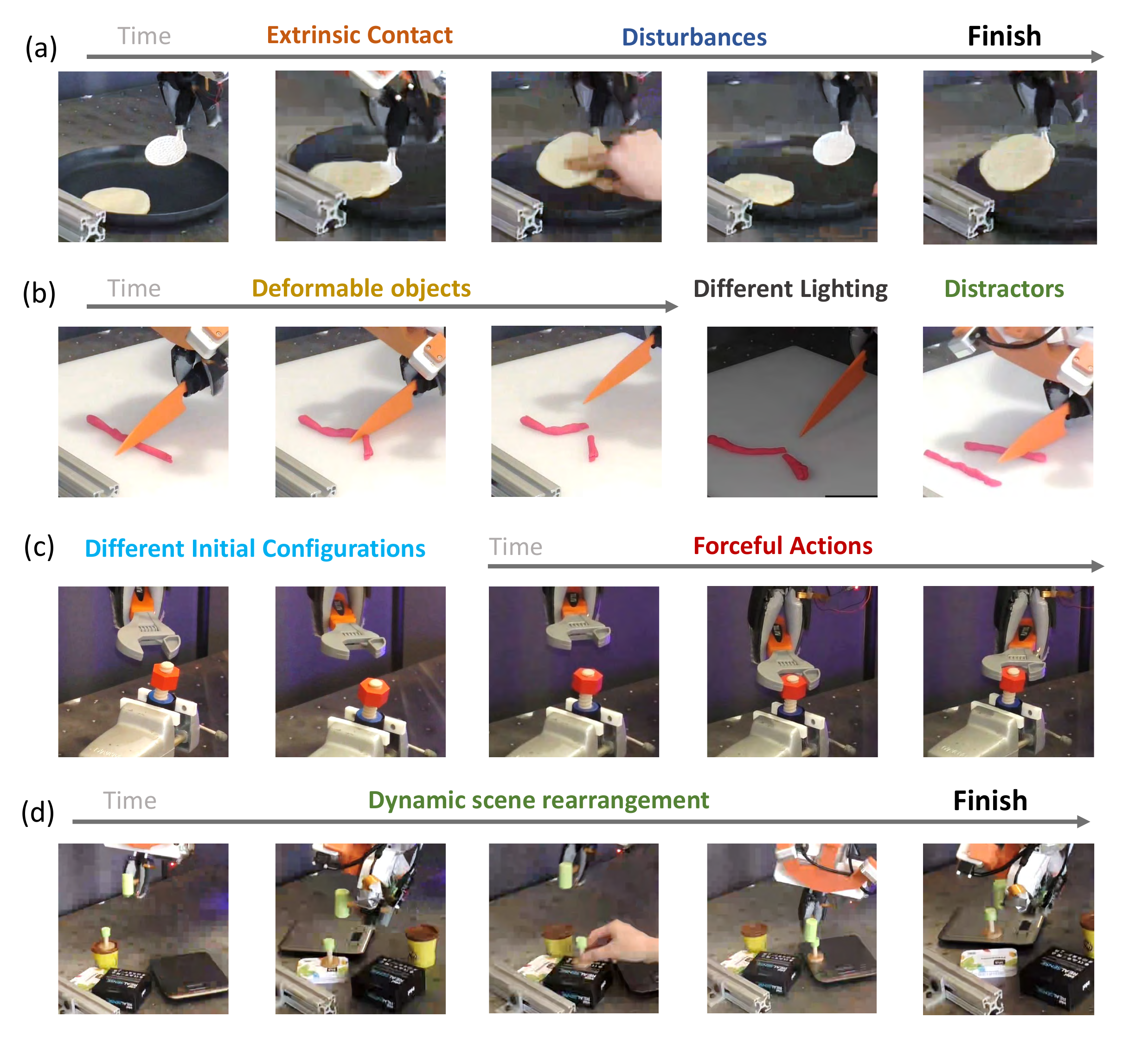}
\vspace{-20pt}
\caption{\textbf{Generalizable Tool Use.} Policy composition can be applied to multiple tool-use tasks using a spatula, knife, wrench, and hammer. Resultant policies generalize across disturbances (a)  and distractors (b) across different initial configurations and settings which require applied force (c) and dynamic rearrangements of scenes (d). The horizontal axis shows the time dimension for each executed trajectory. 
}
\label{fig:realworld_robust}
\vspace{-20pt}
\end{figure}



Similar to how humans learn complex manipulation behaviors that require common sense and reasoning  \cite{osiurak2016tool} through mental simulation, watching, and active interaction, robot policy learning benefits from diverse sources of data. Simulation holds the promise to provide large-scale training data with massive diversity, yet policies trained with simulation usually suffer from the sim-to-real gaps \cite{zhao2020sim}. Real-world human demonstrations are the largest existing data source and they are easy to acquire to teach high-level motions \cite{duan2017one,wang2023mimicplay}. Yet they suffer from embodiment gaps, especially on contact-rich motions. Robot teleop data has the least domain shifts and delivers impressive demos \cite{chi2023diffusion,zhao2023learning}, yet it can be expensive to acquire at large scales. Moreover, data of different modalities is often gathered with different hardware for each task. Given such massive heterogeneity in data, it remains unclear how to train policies on such data, that can achieve both dexterity and {generalization} to new scenes and tasks. 

\begin{figure*}[!t]
\centering 
\includegraphics[width=0.95\textwidth]{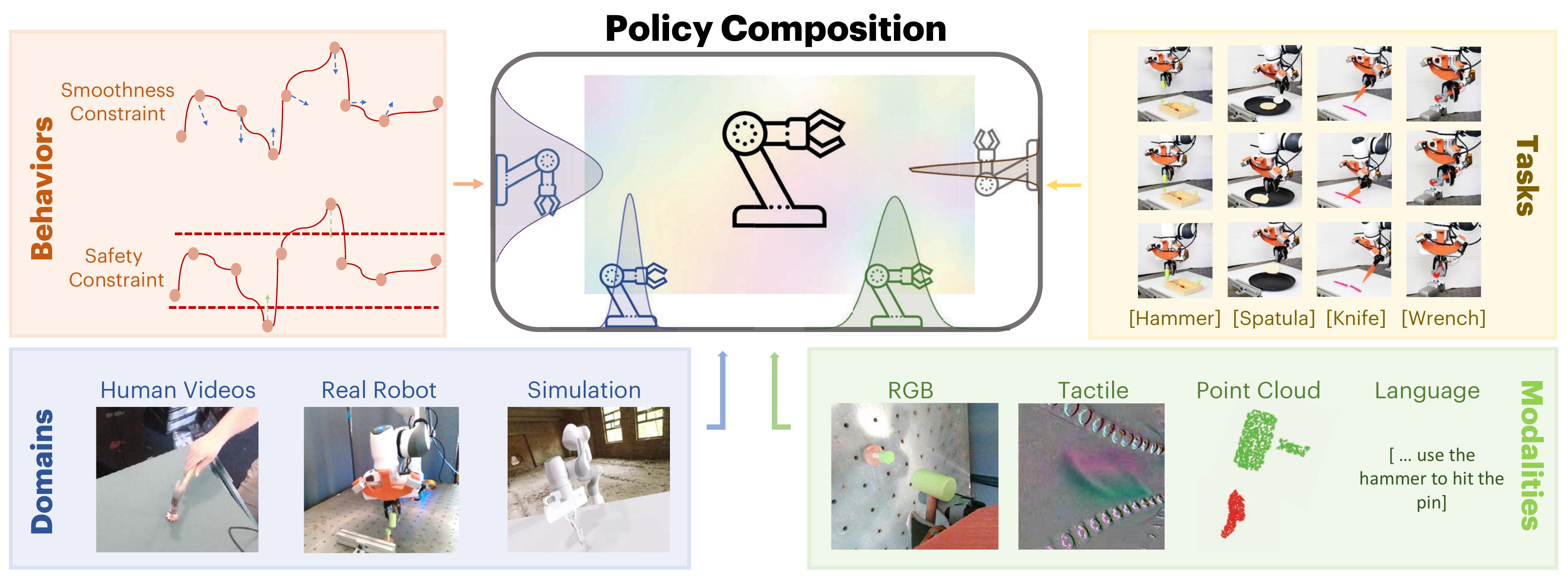}
\vspace{-5pt}
\caption{\textbf{Policy Composition.} \METHOD{} combines information across behaviors, tasks, modalities, and domains using probabilistic composition. This allows our approach to modularly combine information at prediction time (with no retraining), enabling generalization to challenging tool-use tasks, by leveraging information from multiple domains. }
\label{fig:framework}
\vspace{-15pt}
\end{figure*}
We propose Policy Composition (\METHOD), a framework {(Figure \ref{fig:framework})} to \textbf{compose} multiple sources of data and objectives in different modalities and tasks with associated behaviors. This allows us to naturally divide and conquer the combined challenging problem with multiple policies and allow adaptation at test time. Specifically, each policy is instantiated probabilistically using a trajectory level diffusion model~\citep{janner2022planning,ajay2022conditional,chi2023diffusion}, enabling policies to be composed together by {combining score predictions~\citep{urain2023se,reuss2023goal,
hansel2023hierarchical,gkanatsios2023energy,liu2022compositional,du2023reduce}.} {In addition to representing cost functions \cite{urain2023se}, constraints \cite{gkanatsios2023energy,yang2023compositional}, and goals \cite{reuss2023goal}, learned probability distributions can represent policies operating across distinct sensor modalities from domains such as simulation, real robots, and human videos. We apply our methods for the tasks of \textbf{generalizable robotic tool-use}~\cite{ren2023leveraging,qin2020keto,wang2023fleet,seita2023toolflownet} policies}.  We illustrate how different compositions of policies at the task, behavior, and policy levels enable robust generalization to novel scenes and tasks without fine-tuning.

In comparison to approaches to learning generalist policies through representation learning \cite{nair2022r3m} or through large-scale data-pooling \cite{open_x_embodiment_rt_x_2023}, our approach does not require extensive data engineering or data sharing to align observation and action spaces as individual policies are learned modularly on separate domains of data. Furthermore, our modular approach enables us to quickly adapt to out-of-distribution settings with additional sources of data or tasks with control \cite{reuss2023goal,hansel2023hierarchical}, by simply training additional policies without discarding information about prior tasks. In contrast to individually trained components (such as perception subsystem) in a robotic system \cite{brooks1991intelligence}, PoCo allows multiple instances of the policies trained independently to combined in arbitrary combinations in parallel 
to improve performance.  



Via simulation testing on the Fleet-Tools benchmark \cite{wang2023fleet}, we demonstrate that the task composition improves diffusion policies in multi-task settings with various tools such as spatulas and wrenches. We further show that test-time behavior compositions can improve desired behavior objectives such as smoothness and workspace constraints. In the real world {(Figure \ref{fig:realworld_robust})}, we compose across policies trained with different modalities such as image, point cloud, and tactile image, and across domains such as in simulation and the real world. This allows the policies to stay robust across changes in manipulands, distractor objects, camera viewpoints and angles, as well as demonstrate dexterous retrying and fine-grained behaviors. We demonstrate that the overall compositions provide success rate improvements of $20\%$ in both simulation and the real world compared to baselines.

The contributions are summarized as follows: (a)  we propose \METHOD, a policy composition framework to use probabilistic composition of diffusion models to combine information from different domains and modalities (b) we develop task-level, behavior-level, and domain-level prediction-time compositions for constructing complex composite policies without retraining (c) we illustrate the scene-level and task-level generalization of PoCo across simulation and real-world settings for 4 different tool-use tasks, and demonstrate robust and dexterous behaviors.
 
\section{Related Works} 
\label{sec:related_works}

\subsection{Diffusion Models}
Diffusion models \cite{sohl2015deep,ho2020denoising} are a class of probabilistic generative models that iteratively refine randomly sampled noise into draws of underlying distributions.  Compared to other generative models such as Variational Autoencoder (VAE) \cite{kingma2019introduction} and Generative Adversarial Networks (GAN) \cite{creswell2018generative}, diffusion models, or more generally Energy-Based Models \cite{lecun2006tutorial,du2019implicit}, learn the gradient fields of an implicit function which can be optimized at inference time to readily compose with multiple models. Moreover, diffusion model composition has achieved impressive results in generating high-dimensional distributions such as images and videos \cite{singer2022make,ho2022imagen,du2020compositional}.

Recently, diffusion models have been applied to robotic applications \cite{huang2023diffusion,janner2022planning,ajay2022conditional,chi2023diffusion}, and enjoy rigorous theoretical guarantees \cite{block2023provable}. 
These successes are attributable in part to the connection between diffusion processes and traditional gradient-based trajectory optimization, as practiced in robotic planning, optimal control, and reinforcement learning \cite{toussaint2009robot,mukadam2018continuous,levine2018reinforcement,janner2022planning}. 
{The implicit function representation of diffusion models allows them to be composed with external probability distributions~\cite{janner2022planning, dhariwal2021diffusion}, and researchers have had empirical successes in composing multiple diffusion models~\cite{du2020compositional, ho2022classifier, liu2022compositional,nie2021controllable,du2023reduce, yang2023compositional}. }  

\vspace{-5pt}
{
\subsection{Compositional Models in Robotics}
In robotics, probabilistic composition has been previously explored as an approach for structured generalization \cite{du2019model,janner2022planning,gkanatsios2023energy, yang2023compositional, reuss2023goal,mishra2023generative, hansel2023hierarchical,urain2023se} at inference time. Previous works have also investigated a probabilistic view on composing energy-based reactive policies for robotic motion planning such as HiPBI \cite{hansel2023hierarchical}.  In BESO\cite{reuss2023goal},  score functions for goal-conditioned imitation learning are composed using classifier-free guidance of an unconditional and conditional goal policies.  In SREM and CCSP \cite{gkanatsios2023energy,yang2023compositional}, scene rearrangement is accomplished by composing EBMs to represent different subgoals in compositions of object scenes. In SE3 Diffusion Fields \cite{urain2023se}, learned cost functions are used as gradients for joint motion and grasping planning. Leveraging a similar score combination process, our work explores how such probabilistic composition can be applied to learned policies, enabling the combination of multidomain and multitask data to effectively solve complex tool-use tasks.
}


\subsection{Multi-task and Multi-domain Imitation Learning}
Multi-task learning and models that exhibit multi-task behaviors have shown impressive performance in computer vision \cite{zhang2018overview,standley2020tasks}, natural language processing \cite{radford2019language,collobert2008unified}. Some of the most promising approaches in robot learning use direct data pooling, either from robot-only data such as RT series \cite{brohan2022rt,shridhar2023perceiver,octo_2023}, or mixtures of human and simulation data. \cite{,wang2023mimicplay,wang2023gensim,shaw2023videodex}. At the same time,  there is a rich tradition of explicitly accounting for problem structure in multi-task learning \cite{ruder2017overview}, meta-learning \cite{vilalta2002perspective,nichol2018first,finn2017model}, and few-shot learning domains \cite{wang2020generalizing}. Whereas data-pooling approaches require a careful balance of constituent data distributions, and approaches using problem structure \cite{open_x_embodiment_rt_x_2023} require additional algorithmic complication,   our method composes diffusion policies learned from different domains at inference time with minimal algorithmic overhead. 

\subsection{Robotic Tool-Use}
Tool-use is a widely studied problem for robotics \cite{toussaint2018differentiable,holladay2019force} and the AI community \cite{wimpenny2009cognitive,osiurak2016tool,allen2020rapid}. Compared to the well-studied grasping and planar-pushing problems, tool-use tasks require additional skills such as fine contact-rich dexterity, object-object affordance, and compositional generalization  
\cite{osiurak2016tool,wang2023fleet}. To improve tool-use capability of robots, prior works have studied using model-based methods \cite{toussaint2018differentiable,holladay2019force}, sparse perception features \cite{gajewski2019adapting,qin2020keto}, language conditioning \cite{ren2023leveraging,xu2023creative}, simulation and online videos \cite{zorina2021learning}, point cloud inputs \cite{seita2023toolflownet}. Due to the partial occlusion and the extrinsic contact sensing problems during tool-using, some works \cite{sipos2023multiscope,wi2022virdo,suh2022seed,zhao2023gelsight} have studied using tactile sensors for tool manipulations. Our work proposes a novel multi-modal learning system for real robots, humans, and simulations in common tool-use tasks \cite{wang2023fleet}  with industrial and household applications. 
\section{Diffusion Models for Trajectory Generation}

We study how to combine information across different tasks and domains such as simulation environments, human demonstration videos, on-robot data in combination to solve new tasks in new domains. Specifically, following \cite{janner2022planning,chi2023diffusion}, we can parameterize a  generative policy $\pi$ on an action trajectory  $\mathcal{\tau}$ with a conditional distribution $\pi(\tau|o)$ given a history of observations, denoted as $o$, which belongs to a certain set of observation spaces, e.g. the spaces of point clouds, images, or images of tactile sensor deformations.  Regardless of observation space,  $\tau \in \mathbb{R}^{H\times d}$ corresponds to action trajectory at a horizon $H$ and action dimension $d$. Because our action trajectories lie in the same space, even though observations do not, we will be able to compose across multiple different observation spaces as described in what follows.







Restricted to any given observation space $\mathcal{O}_k$, we represent a policy $\pi(\tau|o)$ with a diffusion model, where the trajectory $\tau$ is generated from the policy using an iterative denoising process.
To generate a trajectory from the diffusion model, a noisy sample $\tau_T$ is initialized from Gaussian noise $\mathcal{N}(0, I)$. This sample $\tau_T$ is then iteratively denoised using a learned denoising network $\epsilon_\theta$ as  follows:
\begin{equation}
    \tau^{t-1}=\tau^{t}-\gamma_t \epsilon_\theta(\tau^t,t \mid o) + \xi, \quad \xi \sim \mathcal{N} \bigl(0, \sigma^2_t I \bigl),
    \label{eq:unconditional_langevin}
\end{equation}
for $T$ total steps of denoising with specified per-timestep noise magnitudes $\sigma_t$ and step size $\gamma_t$. The final generated sample $\tau_0$ corresponds to the predicted trajectory of the policy $\pi_\theta$ 

To train the denoising network $\epsilon_\theta$, we corrupt the trajectory $\tau$ across a randomly selected noise level $t \in \{1\ldots T\}$, and a corresponding corruption  $\epsilon^t = \alpha^t \epsilon$ is added to a clean trajectory $\tau_0$. The denoising model $\epsilon_\theta$ is trained to predict the added noise to the trajectory using the MSE objective:
\begin{equation}
\label{eq:diffusion_loss}
\mathcal{L}_{\text{MSE}}=\|\mathbf{\epsilon}^t - \epsilon_\theta(\tau_0 +  \mathbf{\epsilon}^t, t \mid o)\|^2.
\end{equation} 

The denoising function $\epsilon_\theta$ estimates the score $\nabla_{\tau} E(x)$ of an underlying EBM (unnormalized) probability distribution~\cite{vincent2011connection,liu2022compositional}, where $e^{-E(\tau)}$ represents the desired trajectory distribution $\pi(\tau \mid o)$, and it also allows diffusion models to be easily composed with other probability distributions.




\section{Learning Compositional Policies}
\label{sect:policy_composition}

In this section, we introduce the idea of compositional policy learning to handle the heterogeneous training and testing settings in robotics. For two probability distributions $p_1,p_2$, we compose them in the {following form \cite{hinton2002training,du2020compositional,du2023reduce,mishra2023generative,hansel2023hierarchical,gkanatsios2023energy,urain2023se}}
\begin{equation}
    {p_{\text{product}}(\tau) \propto p_1(\tau)p_2(\tau).}
\end{equation}
{This product distribution combines the information contained in each distribution -- for instance, when $p_1(\tau)$ and $p_2(\tau)$ represent Gaussian PDFs of the uncertainty from two sensor readings, the composed product Gaussian PDF $p_{\text{product}}(\tau)$ combines the mean and uncertainty estimates across both sensors.  To expand this general composition idea in robotic policy learning with heterogeneous data and diffusion models}, we first introduce how policies learned across data domains and modalities of tasks and objectives can be represented as probability distributions in Section~\ref{sect:robot_hetero}. In Section~\ref{sec:domain_comp}, we discuss how we can {compose different probability distributions in the form of a product distribution} to leverage information across different modalities, domains, tasks, and behaviors. Finally, we discuss implementation details of how to sample from composed policies using diffusion models in Section~\ref{sect:implementation}.

\subsection{A Probabilistic Perspective on Robotic Heterogeneity}
\label{sect:robot_hetero}

We propose a {probabilistic perspective \cite{du2020compositional,yang2023compositional,gkanatsios2023energy, mishra2023generative}} to combine information across different domains such as simulation, humans, and real robots, across sensory modalities such as RGB images and point clouds, and across different behavioral constraints such as smoothness and collision avoidance. Training a generic policy directly across all these sources of information has been an outstanding challenge due to these heterogeneous modalities and objectives present in the data.




\begin{figure}[!t]
\centering 
\includegraphics[width=\linewidth]{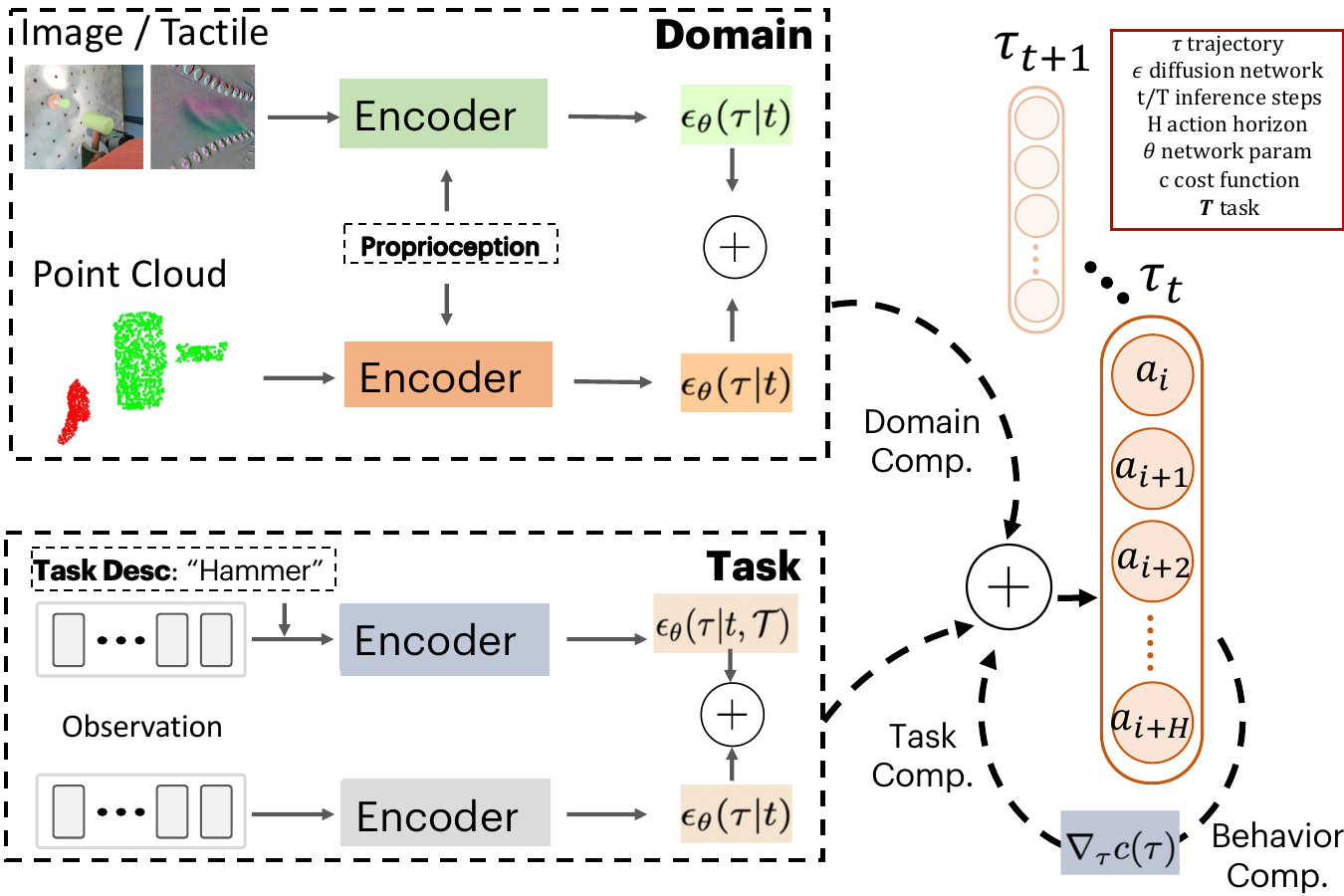}
\caption{\textbf{Illustration of Policy Composition.} Policies are composed at test time by combining gradient predictions. This can apply to \textit{domain composition} by combining policies that train with different modalities such as image, point cloud, and tactile images. It can also be used with different tasks in \textit{task composition} and additional cost functions for desired behavior with \textit{behavior composition}. The only assumption is that the diffused outputs for each model need to be in the same space, i.e. the action dimension and action horizon. We denote the composition operator using a ``plus'' sign.  }
\label{fig:composition_arch}
\vspace{-10pt}
\end{figure}


\newcommand{\cM}{\mathcal{M}}
Let us first describe how we model these different forms of heterogeneity: domain, behavior, task, and sensing modality. We let \emph{modality} $\cM \in  \{\mathcal{M}_{\text{tactile}},\mathcal{M}_{\text{pointcloud}},\mathcal{M}_{\text{image}}, \mathcal{M}_{\text{proprioceptive}}\}$ denote a sensing modality - tactile images, pointcloud, RGB image, or proprioceptive information including joint angles and end effector poses. We describe below how policies can take as input observations from any combination of these modalities.


Next, we consider multiple \emph{data domains} $\mathcal{D}$ of demonstration trajectories.  These include simulation data $\mathcal{D}_{\mathrm{sim}}$,  human video demonstrations $\mathcal{D}_{\mathrm{human}}$, or on-robot teleoperated data saved from robot sensors $\mathcal{D}_{\mathrm{robot}}$. In addition to different data distributions, different domains also have (a) different control frequencies, and (b) data from different combinations of modalities but crucially (c) \emph{must have the same actions spaces} to enable composition in our current approach.


We then allow for the constraints on the desired behavior via a cost function $c(\tau)$ on the action trajectories, such as smoothness costs ${c}_{\text{smoothness}}(\tau)$ and safety costs ${c}_{\text{safety}}(\tau)$. Finally, we allow the specification of the robot task $\mathcal T$ via natural-language text command, such as $\mathcal{T}_{\text{hammer}}$: ``hammering with a hammer'' or $\mathcal{T}_{\text{spatula}}$: ``scooping with a spatula''.

We now propose to learn a separate probabilistic model $p_{\mathcal{D}}^{\mathcal{M}}(\cdot|c,\mathcal{T})$ on each tuple $(\mathcal M, \mathcal D, \mathcal T, c)$ of (modality, domain,  task, behavioral cost). 
We begin by training a policy $p_{D}^{\mathcal{M}}(\cdot \mid  \mathcal{T})$ with no specified cost to exhibit a high likelihood on trajectories that are similar to expert demonstrations to achieve certain tasks $\mathcal{T}$ seen in domain $\mathcal{D}$ and modality $\mathcal{M}$. We instantiate this learned model as a denoising diffusion model.  {At inference time, we can  incorporate the costs by sampling from the EBM distribution $p_{D}^{\mathcal{M}}(\tau \mid  c, \mathcal{T}) \propto \exp(-c(\tau)) \cdot p_{D}^{\mathcal{M}}(\tau \mid \mathcal{T})$. In this setting, we choose to convert each cost probabilistically through the Boltzmann distribution $\exp(-c(\tau))$, although alternative probabilistic parameterizations are also valid as long as cost exhibits high-likelihood when the constraints are satisfied in a trajectory and low-likelihood when they are not.}   Concretely,  our smoothness cost  $c_{\text{smooth}}(\tau)=\norm{\tau''}^2$  penalizes a discrete approximation of the second derivative of the trajectory; and workspace safety constraint cost $c_{\text{safety}}(\tau)=\norm{\min(\tau_{min}-\tau,0)}^2+\norm{\max(\tau -\tau_{max},0)}^2$, where $\tau_{min}, \tau_{max}$ represent certain trajectory bounds.

Importantly, rather than \emph{pooling all the data naively}, we train separate models for different (modality, domain, task) combinations and use compositionality, as described below, as a way to learn from this diverse and heterogeneous data.

\subsection{Combining Information through Policy Composition}
\label{sec:domain_comp}

Given information across trajectories encoded by two probability distributions $p_{\mathcal{D}}^{\mathcal{M}}(\cdot|c,\mathcal{T})$ and $p_{\mathcal{D}'}^{\mathcal{M}'}(\cdot|c',\mathcal{T}')$, {in policy composition, we propose to directly combine the information across both distributions at inference time by sampling from the product distribution. The key idea is to combine the score predictions from diffusion policies at inference time.
 

{Under the notations defined in the previous section, we define}
%
%
%
%
\begin{equation}
\label{eq:product}
    p_{\text{product}} \propto p_{\mathcal{D}}^{\mathcal{M}}(\cdot|c,\mathcal{T}) \cdot p_{\mathcal{D}'}^{\mathcal{M}'}(\cdot|c',\mathcal{T}'),
\end{equation}
 
 Note that we omit the observation $\mathcal{O}$ of the policy in $p$ to further simplify the notation. This $p_{\text{product}}(\tau)$ will exhibit a high likelihood at all trajectories that are valid under both distributions, effectively encoding the information from both probability distributions. This approach allows each component neural network to be trained \textit{individually}, and then \textit{modularly} combined at test time, which further enables the incorporation of new sources of information.

 {Furthermore, as discussed in Appendix~\ref{appendix:proof} this composition obtains the precise form of the conditional distribution given composed tasks and costs under the assumption that (1) there is mutual independence of tasks $\mathcal{T}$ and costs $c$ and (2) conditional independence of tasks $\mathcal{T}$ and costs $c$ given a trajectory $\tau$. These assumptions hold when policies when different constraints and tasks are mutually independent. For instance, policy that is trained to mimic pancake lifting behavior and the costs of maintaining trajectory smoothness satisfy both assumptions.  
  }
 Below, we provide three examples of how policy composition can be used to improve policy performance. 



\noindent \textbf{Task-level Composition.} \label{sec:task_compo} Given an unconditional distribution $p(\tau)$ describing possible trajectories for certain domains and modalities, and a conditional distribution $p(\tau|\mathcal{T})$ describing possible trajectories for a specified task $\mathcal{T}$, we can use composition to combine information across both distributions to synthesize a policy more likely to accomplish task $\mathcal{T}$.

In particular, we can construct the policy composition
\begin{equation}
\label{eq:prob_task_comp}
    p_{\text{product}}(\tau) \propto p(\tau) p(\mathcal{T}|\tau)^\alpha \propto p(\tau) \left (\frac{p(\tau|\mathcal{T})}{p(\tau)} \right)^\alpha,
\end{equation}
for a chosen $\alpha > 1$. This particular policy composition puts extra weight on trajectories that are likely to accomplish task $\mathcal{T}$ (based off the weight $\alpha$), improving the final quality of synthesized trajectories. {This task-level composition procedure does not require separate training for each task, and trains a general policy that can achieve multiple task objectives, resembling a classifier-free diffusion model \cite{ho2022classifier}, {which has also been applied to goal-conditioned policies~\citep{reuss2023goal}} .}


\begin{figure}[!t]
\centering 
\includegraphics[width=1\linewidth]{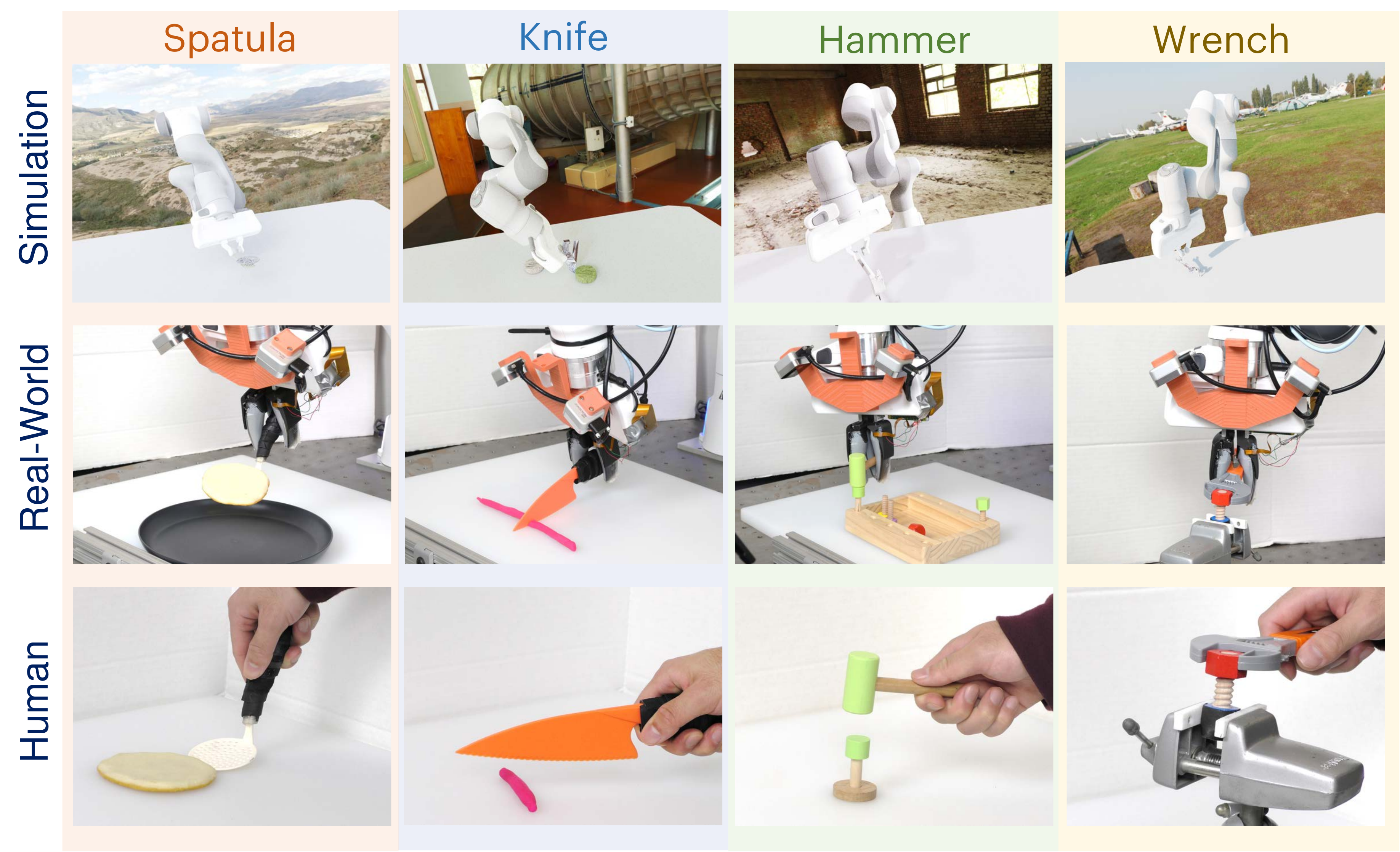}
\caption{\textbf{Task Visualizations.} Visualization of Tool-Use Tasks, ``spatula, knife, hammer, wrench'' are studied in this work in three different domains: simulation, real-world robots, and human demonstrations. }
\label{fig:alltasks}
\vspace{-10pt}
\end{figure}

\begin{figure*}[!t]
\centering 
\includegraphics[width=0.99\linewidth]{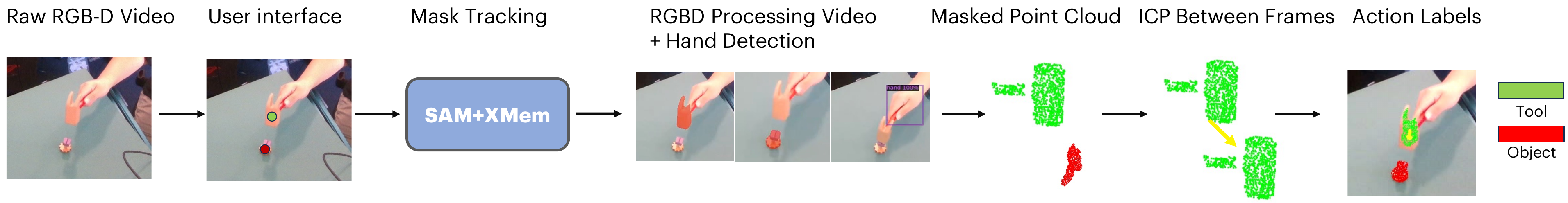}
\vspace{-5pt}
\caption{\textbf{Dataset Processing Pipeline.} We show the human demonstration video collected in the wild from uncalibrated RGB-D cameras can be converted into a labeled trajectory (relative tool poses). The same pipeline applies to evaluating policy in the real world. }
\vspace{-15pt}
\label{fig:data_processing}
\end{figure*}

\begin{figure}[t]
\centering
\small
\begin{minipage}{\linewidth}
\begin{algorithm}[H]
\caption{Sampling Process in Policy Composition}
\label{algo:composition} 
\hrule
\begin{algorithmic}[1] 
\STATE \textbf{Input}: Tasks $\{\mathcal{T}_k\}_{k=1}^K$, Policies Per Domain $\{\theta_i\}_{i=1}^N$, Costs $\{c_j(\tau)\}_{j=1}^M$, Base Multitask Policy $\phi$. 
\STATE {\bf Init}: $\tau_0$ from Standard Gaussian 
\FOR{{Iteration $t=0\dots,T-1$}}
 \STATE \textbf{Initialize denoising step: $\nabla{\tau^t}=0$}
\FOR{{Task $k=1\dots,K$}}
 \STATE \textbf{Task Composition:} $\nabla{\tau^t}= \nabla{\tau^t}+\epsilon_\phi(\tau|t) + \alpha (\epsilon_\phi(\tau|t, \mathcal{T}_k) - \epsilon_\phi(\tau|t))$ \\
 \ENDFOR 
 \FOR{{Behavior $j=1\dots,M$}}
 \STATE \textbf{Behavior Composition:}  $\nabla \tau^t=\nabla{\tau^t}+\gamma_{{c}} \nabla_{\tau}c_j(\tau)$,  \\
  \ENDFOR 
 \FOR{{Domain $i=1\dots,N$}}
 \STATE \textbf{Domain Composition:} $\nabla \tau^t=\nabla{\tau^t}+\gamma_{\mathcal{D}} \epsilon_{\theta_i}(\tau|t)
  $
\ENDFOR 
\STATE \textbf{Diffusion Step:} $\tau^{t+1}=\tau^t-\gamma \nabla \tau^t+\xi$
 \ENDFOR 
\vspace{.2em}
\STATE {\bf Return}: denoised trajectory $\tau_0$ 
\end{algorithmic}
\end{algorithm}
\end{minipage}
\vspace{-15pt}
\end{figure}

\begin{figure*}[!t]
\centering 
\includegraphics[width=1.\linewidth]{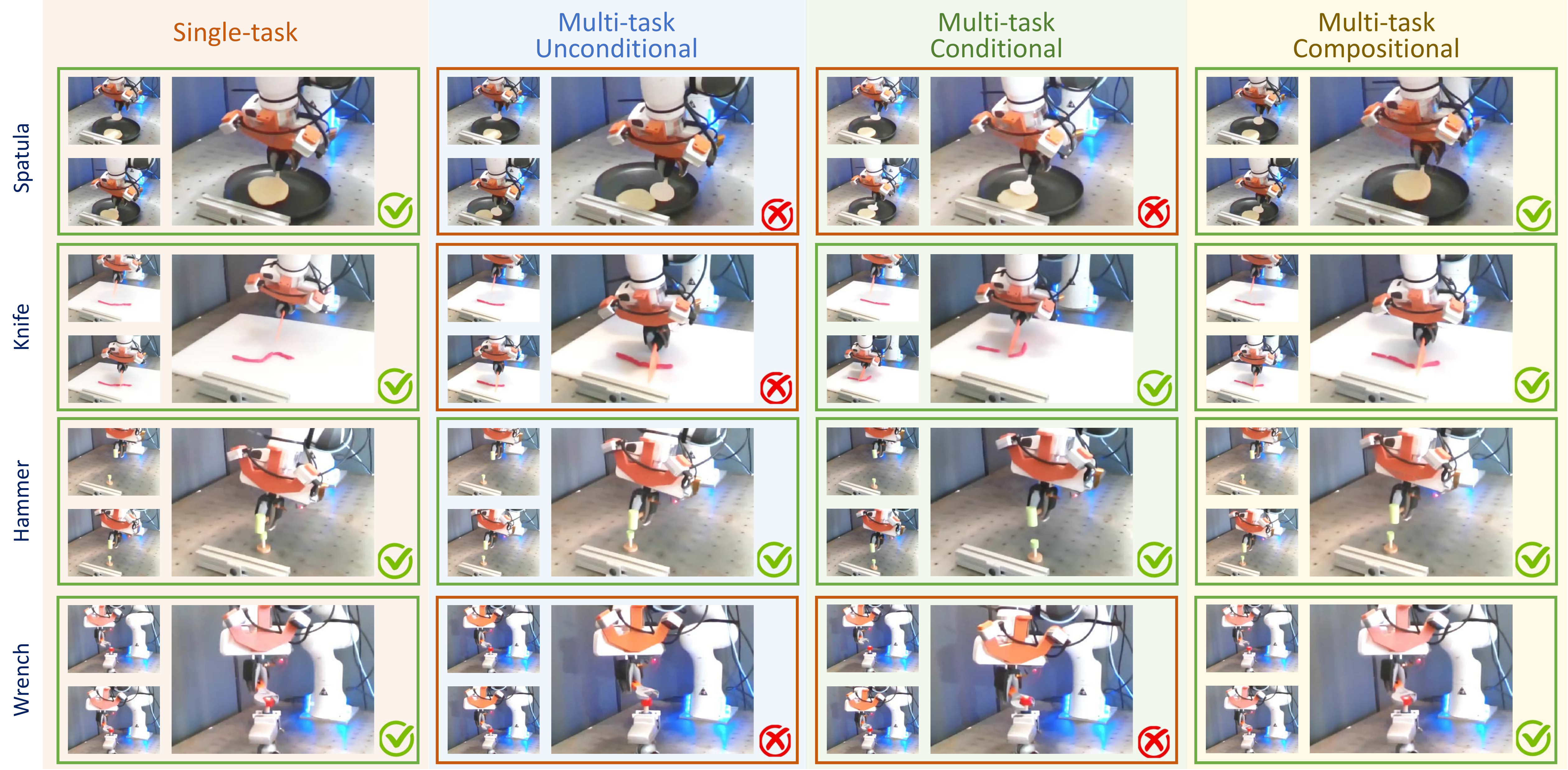}
\caption{\textbf{Task-level Qualitative Results.} By composing unconditional and task-conditional models, our composed policy can accomplish a diverse set of tasks and outperform unconditioned multitask policies. }
\label{fig:task_gen}
\vspace{-12pt}
\end{figure*}

\noindent \textbf{Behavior-level Composition.} \label{sec:behavior_compo} 
Given a distribution $p(\tau|c,\mathcal{T})$ of trajectories for a task $\mathcal{T}$ and a distribution $p_{\text{cost}}(\tau|c)$ encoding a cost function over a trajectory, we can combine distributions using
\begin{equation}
    p_{\text{product}}(\tau) \propto p_{\text{cost}}(\tau) p(\tau|\mathcal{T}).
\end{equation}
This policy composition combines information across the task distribution and the cost objective, ensuring that synthesized trajectories both accomplish a task and optimize a specified cost objective. {This inference procedure is flexible and is similar to what a robotic trajectory optimizer will be able to do}. In the action diffusion setting, the action trajectories need to be integrated to get the state trajectories.



\noindent \textbf{Domain-level composition.} \label{sec:policy_compo} Given two policy distributions $p_{D_1}(\tau|\mathcal{T}),p_{D_2}(\tau|\mathcal{T})$ representing information about valid trajectories to solve a task $\mathcal{T}$ trained on different sensor modalities and domains $D_1,D_2$, we can combine the information across both models together through
\begin{equation}
    p_{\text{product}}(\tau) \propto p_{D_1}(\tau|\mathcal{T}) p_{D_2}(\tau|\mathcal{T}).
\end{equation}
This policy composition allows us to leverage information captured from different sensor modalities and domains, which is useful for complementing strengths of data gathered in one domain with another, {\it i.e.} when real-robot demonstrations are expensive to gather but more accurate, compared to simulation demonstrations which are cheaper to gather but not as accurate. We use feature concatenation for different modalities in the same domain for simplicity. Finally, note that multiple compositions may be nested together to further combine information across policies.

\subsection{Implementing Compositional Sampling}
\label{sect:implementation}

One way of viewing the policy composition is through energy-based models (EBM). Given two EBM distributions~\citep{du2019implicit} $p_1(\tau) \propto e^{-E_1(\tau)}$ and $p_2(\tau) \propto e^{-E_2(\tau)}$, the product of the two distributions $p_1(\tau)p_2(\tau)$ can be represented as a EBM $e^{-(E_1(\tau) + E_2(\tau))}$. We can sample from an EBM distribution $e^{-E(\tau)}$ using Langevin dynamics where given a sample initialized from $\tau^0$ initialized from $\mathcal{N}(0, 1)$ we iteratively refine it using the gradient of the energy function
\begin{equation}
    \tau^{t-1}=\tau^{t}-\gamma \nabla_{\tau}E(\tau) + \xi, \quad \xi \sim \mathcal{N} \bigl(0, \sigma^2_t I \bigl),
    \label{eq:unconditional_ebm_langevin}
\end{equation}
where $\gamma$ and $\sigma_t$ are hyperparameters in realizing accurate sampling from the probability distribution. The Langevin sampling procedure in Equation~\ref{eq:unconditional_ebm_langevin} corresponds to the typical sampling procedure in diffusion models in Equation~\ref{eq:unconditional_langevin}. 

To combine policies, we compute $\nabla_\tau E_\theta(\tau)$ for each diffusion policy in each domain. Then, by using the score prediction of the denoising network $\epsilon_\theta(\tau, t)$ for each cost function, we can explicitly compute $\nabla_\tau c(\tau)$. This then allows us to use the standard diffusion sampling procedure in Equation~\ref{eq:unconditional_langevin} with the denoising network prediction substituted with the gradient prediction corresponding to composed distributions. 

Concretely, this enables us to sample from task-level composition by using the sampling expression
\begin{equation}
\label{eq:grad_task_comp}
    \tau^{t-1}=\tau^{t}-\gamma_{\mathcal{T}} (\epsilon_\theta(\tau|t) + \alpha (\epsilon_\theta(\tau|t, \mathcal{T}) - \epsilon_\theta(\tau|t)) ) + \xi,
\end{equation}
where $\xi \sim \mathcal{N} \bigl(0, \sigma^2_t I \bigl)$. These two networks $\theta(\tau|t, \mathcal{T}),\theta(\tau|t)$ can be trained jointly with classifier-free training \cite{ho2022classifier}. 
This  enables us to sample from the behavior-level composition by using the sampling expression 
\begin{equation}
    \tau^{t-1}=\tau^{t}-\gamma_c (\epsilon_\theta(\tau|t, \mathcal{T}) + \nabla_{\tau}c(\tau)) + \xi, \
    \label{eq:objective_level}
\end{equation}
where $\xi \sim \mathcal{N} \bigl(0, \sigma^2_t I \bigl)$.
Finally, assuming we have trained different networks $\theta_1,\theta_2$ for different domains $\mathcal{D}_1,\mathcal{D}_2$, we can sample from domain-level composition by using the sampling expression
\begin{equation}
    \tau^{t-1}=\tau^{t}-\gamma_{\mathcal{D}} (\epsilon_\theta^1(\tau|t ) +\epsilon_\theta^2(\tau|t )) + \xi,
    \label{eq:policy_level}
\end{equation}
where $\xi \sim \mathcal{N} \bigl(0, \sigma^2_t I \bigl)$.


We note again that different compositions can use different observation inputs: for example, behavior composition does not rely on external observation, and domain composition can compose two separate policies using point cloud or image input respectively. In practice, we tuned the composition weights $\gamma$ to demonstrate the composition behavior. {Since we assume all models have shared the same action bounds for output normalization, the gradients (or noise prediction) with respect to the unnormalized trajectory (such as analytic costs) can be approximatedly combined with the normalized ones with a fixed linear transform and a scalar weight.} It is also possible to use more advanced sampling techniques \cite{du2023reduce} and nested composition with an ensemble or mixture of policies for more complex behaviors. See Algorithm \ref{algo:composition} and Figure \ref{fig:composition_arch} for a summary of the sampling process.

\begin{figure*}[!t]
\centering 
\includegraphics[width=\linewidth]{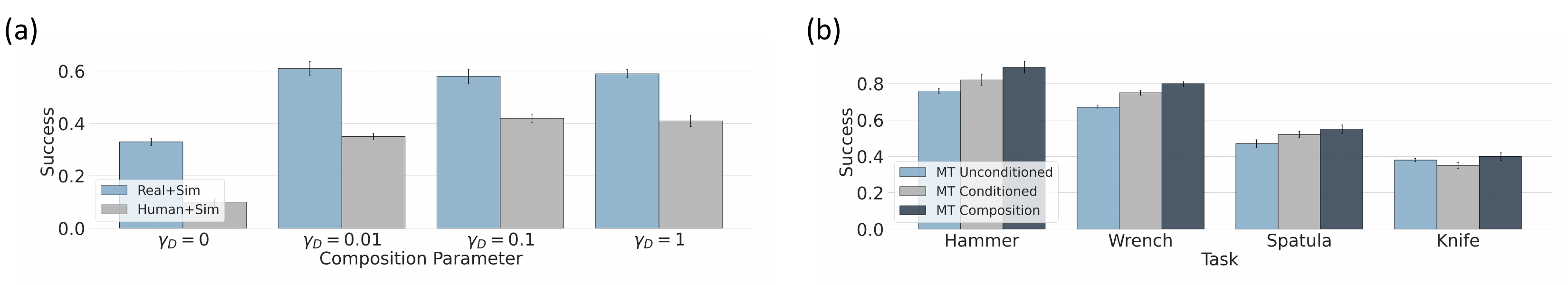}
\vspace{-10pt}
\caption{\textbf{Effect of Policy Composition in Simulation.} \textbf{(a)} Task Composition performs the best in multitask policy evaluation in simulation. \textbf{(b)} Simulation policies help with composition across domains and evaluation in simulation. }
\label{fig:sim_res}
\vspace{-3pt}
\end{figure*}

\section{Implementations}
 We use a temporal U-Net structure with Denoising Diffusion Probabilistic Model (DDPM) \cite{chi2023diffusion,janner2022planning} at training time for 100 steps and Denoising Diffusion Implicit Model (DDIM) at testing time for 32 steps \cite{chi2023diffusion,janner2022planning}.  Since the goal is to compose different diffusion models across domains $\mathcal{D}$ and tasks $\mathcal{T}$, we use the same action space for all models. For simplicity, a fixed normalization was used for the action bounds rather than dataset statistics, and the robot was controlled using end-effector velocity control.  We focus on \textit{generalizable tool-use tasks} to evaluate the proposed framework.  The task is determined successful when certain thresholds are met. For example, a hammering task is deemed successful when the pin is hammered down.
 We use four task families (wrench, hammer, spatula, and wrench) for evaluation. We use ResNet-18 \cite{he2016deep} as the image encoder and PointNet \cite{qi2017pointnet} as the point cloud encoder. See Figure \ref{fig:alltasks} and Appendix \ref{appendix:dataset} for more detailed discussion of different \textit{domains}. 

\subsection{Simulation Setup}
In the Fleet-Tools benchmark \cite{wang2023fleet}, we use the segmented tool and object point clouds as the agent observations, and the 6-DoF translation and rotation at the end-effector frame as the agent actions. The episode length is around 40 steps and the control frequency is 6Hz. 
Initial end-effector poses, initial object poses, and the tool-in-hand poses are randomized at each episode.  
In total, we collected around 50,000 simulated data points for each tool-object pair across 4 tool families. 
The demonstration trajectories in simulation are calculated with keypoint-based trajectory optimization. We apply data augmentation on both point clouds and actions in the training process. The test scenes are held out from the training process. {We apply data augmentations that maintain the object-tool relationships. 
We also add point-wise noises, random cropping, and random dropping to the observed $512$ tool and $512$ object point clouds from the depth images and masks.  Note that in the simulation experiment, only the point clouds are used for experiments (for their speed and ease of training). In the real-world experiments below, RGB-based policies are used for real-world collected data. }
See Appendix  \ref{appendix:impl} for more details.

\subsection{Real Robot Setup}
We mount a wrist camera (Intel D405) and an overhead camera (Intel D435) to obtain local and global views of the scene. The near view range of D405 is useful for capturing the tool point clouds from the wrist view. We used a simplified version of the GelSight Svelte Hand \cite{zhao2023gelsight,zhao2023gelsight2} installed on a Franka Emika Panda robot to handle different tools. One of the three GelSight Svelte tactile fingers is activated, which provides rich tactile information on tool poses, tool shape, as well as tool-object contacts. For teleoperation, we use an Oculus Quest Pro to track joystick poses, which are mapped to the velocity of the end-effector as action labels for imitation learning. The episode lengths of real-world teleoperation vary from 20 steps to 100 steps. The real robot policy frequency is 10 Hz when using RGB inputs only, or 6 Hz when segmented point clouds are used.  See Appendix  \ref{appendix:dataset} for more details.

\subsection{Human Demonstration Setup}
The demonstration videos can be collected from uncalibrated cameras in the wild. The videos can then be labeled by a few clicks for only the first frame in each trajectory. We then use pre-trained segment-anything \cite{kirillov2023segment} for segmentation, XMem \cite{cheng2022xmem} for tracking, and hand detection pipelines \cite{shan2020understanding} to extract an approximate hand pose, as shown in Figure \ref{fig:data_processing}. 
Point clouds of the tool and the object are extracted and transformed to the hand frame.
Then we use the Iterative Closest Point (ICP) method to solve for the relative movements of the tool between two timesteps, which are used as action labels. 
Trajectory lengths vary from 20 steps to 100 steps. See Appendix  \ref{appendix:dataset} for more implementation details. Additionally, we plan to open-source all collected datasets for future research.

\section{Simulation Experiments} 
\label{sec:sim_experiments}

\begin{table}
\centering
\small
\begin{minipage}[t]{.95\linewidth}
\setlength{\tabcolsep}{4pt}
{
\begin{tabular}{l|ccc}
{\bf Metric} & {\bf Success $\uparrow$} & {\bf Smoothness $\downarrow$} & {\bf Workspace $\downarrow$} \\ 
\midrule
Normal & $0.70$ & 0.027   & 0.030     \\
+Smoothness &0.67 & \textbf{0.016} & 0.038   \\
+Workspace & 0.67 & 0.019 & \textbf{0.022}  \\
\bottomrule
\end{tabular}
}
\end{minipage}
\caption{\textbf{Effect of Behavior Composition.} By combining costs probabilistically, we can optimize metrics from each cost objective.}
\label{tab:behavior_comp}
\vspace{-20pt}
\end{table}

In this section, we explore the following question: when the test distribution is different from any single one of the training distributions, how each compositionality as described in Section \ref{sect:policy_composition} can help adjust policy behaviors (\ref{sec:eval_behavior_compo}), generalize across multiple tasks (\ref{sec:eval_task_compo}) and multiple domains (\ref{sec:eval_domain_comp}) without retraining? 
In simulated experiments, we use the masked tool-object point clouds as the policy input $\mathcal{O}$. $\mathcal{O}$ is further encoded with a PointNet \cite{qi2017pointnet} before being fed into the diffusion model $\epsilon_\theta$. For each task, we perform 10 independent runs of each experiment on 50 scenes, and the average success rates together with its success criteria are reported.  The model size and dataset size are the same for each domain in each experiment across baselines.
The simulation environment is illustrated in Figure \ref{fig:sim_tasks} and is used to evaluate both policies trained from real world data and simulation data.

\begin{table*}

\begin{minipage}[t]{\linewidth}
\setlength{\tabcolsep}{5pt}
\centering
\small
\begin{tabular}{l|ccccc}
& &  & &  {\bf Composition} &   {\bf Composition}   \\ 
{\bf Setting}   &
  {\bf Human} &  {\bf Simulation} & {\bf Real-Robot} &   {\bf (Human+Real)}  &   {\bf (Sim+Real)}    \\ 
  \midrule
Vary Object Pose &  1/5   &  5/5   & 2/5   & 4/5 &  5/5    \\
Vary Robot and Tool Pose &  1/5   & 5/5    & 4/5   & 5/5   & 5/5    \\
Distractor &  0/5   & 5/5    & 2/5   & 3/5   &  5/5   \\
Novel Instance &  1/5   &  3/5   & 2/5   &  1/5  &  5/5   \\
\midrule
Total &   {$15\pm 2.2\%$}   & {$90\pm 4.3\%$}    &   {$50\pm 4.3\%$}  &  {$65\pm 7.4\%$} & {$100\pm 0\%$}   \\
\bottomrule
\end{tabular}
\end{minipage}
\caption{\textbf{Quantitative Results on Domain Composition.} Policy Composition improves {average success rates} compared to individual constituent policy across different scenes on four separate generalization axes, evaluated on the hammering task. }
\vspace{-15pt}
\label{tab:scene_gen}

\end{table*}
\subsection{Behavior-level Composition}
\label{sec:eval_behavior_compo}
In Table \ref{tab:behavior_comp}, we use test-time inference to compose behaviors such as smoothness and workspace constraints.  
Recall that the smoothness costs $c_{\text{smooth}}$ measure the squared norm of the averaged accelerations of the rollout trajectories, where workspace costs $c_{\text{safety}}$ measure the squared norm of the trajectory outside of the workspace bounds. For each environment timestep, we integrate the action trajectories into the workspace pose trajectories and then compute the smoothness and collision costs, and compute its gradients with pytorch autograd for composition. We fix the composition weight to be $\gamma_{c}=0.1$ in this setting.  This maps to a combined diffusion step with the analytic cost functions $\epsilon^t=\epsilon_{\theta}(\tau)+\gamma_{{c}}c(\tau)$ in each step of the diffusion process. In Table \ref{tab:behavior_comp}, we demonstrated the gains in smoothness and safety constraints by adding prior information to the action trajectories in the inference process. This composition is analogous to trajectory optimization widely used in robotic motion planning.

\subsection{Task-level Composition}
\label{sec:eval_task_compo}

In Figure \ref{fig:sim_res} (b), we show that the task composition of conditional and unconditional multi-task tool-use policies outperforms both unconditional and task conditional multi-task tool-use diffusion policies. To construct a conditional multi-task tool-use policy, we assign each tool-use task a text name, such as ``wrench'', and train a conditional tool-use policy based off this text by using a T5 encoder \cite{ni2021sentence} corresponding to learning 
 $\epsilon_\theta(\tau|\mathcal{T})$. To construct an unconditional multi-task tool-use policy, we dropout text labels with a frequency $0.1$ when training our conditional multi-task tool-use policy and replace the text embedding with a fixed latent vector corresponding to learning $\epsilon_\theta(\tau)$.  Our task-level composition then corresponds to $(\epsilon_\theta(\tau) + \alpha (\epsilon_\theta(\tau|\mathcal{T})-\epsilon_\theta(\tau))$, which can also be seen as  classifier-free guidance \cite{ho2022classifier}. From Equation  \ref{eq:prob_task_comp} and its gradient form \ref{eq:grad_task_comp}, we see that when the task weight $\alpha=0$, this policy maps to unconditioned multitask policies, when $\alpha=1$, it maps to standard task-conditioned policies, and when $0<\alpha<1$, we are interpolating between task conditioned and task unconditional policies. When $\alpha>1$, we can get trajectories that are more task-conditioned.  We fix the composition hyperparameter $\alpha=1.5$ in Figure \ref{fig:sim_res} and illustrate how such an approach improves performance over both unconditional and conditional policies.

\begin{figure}[t]
\centering 
\includegraphics[width=\linewidth]{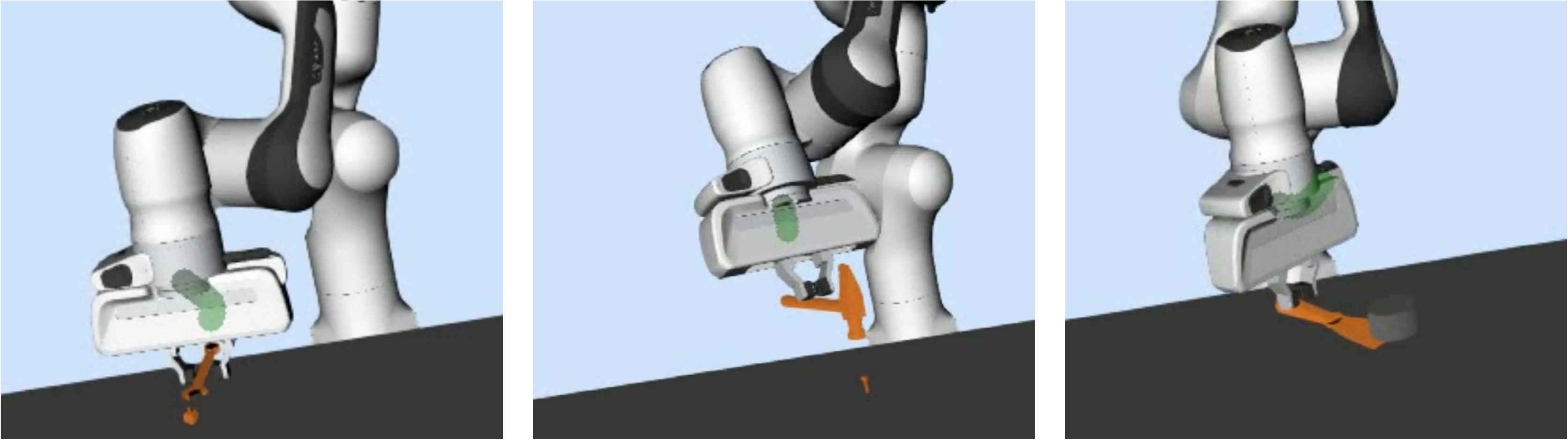}
\caption{\textbf{Simulation Qualitative Results.} We show some visualization with the inferred trajectory (in green) in the Drake simulation environments. These are tool-use tasks using a wrench, hammer, and spatula (left, middle, right). The tools are welded to the end effector.}
\label{fig:sim_tasks}
\vspace{-5pt}
\end{figure}

\begin{figure}[t]
\centering 
\vspace{-0pt}
\includegraphics[width=\linewidth]{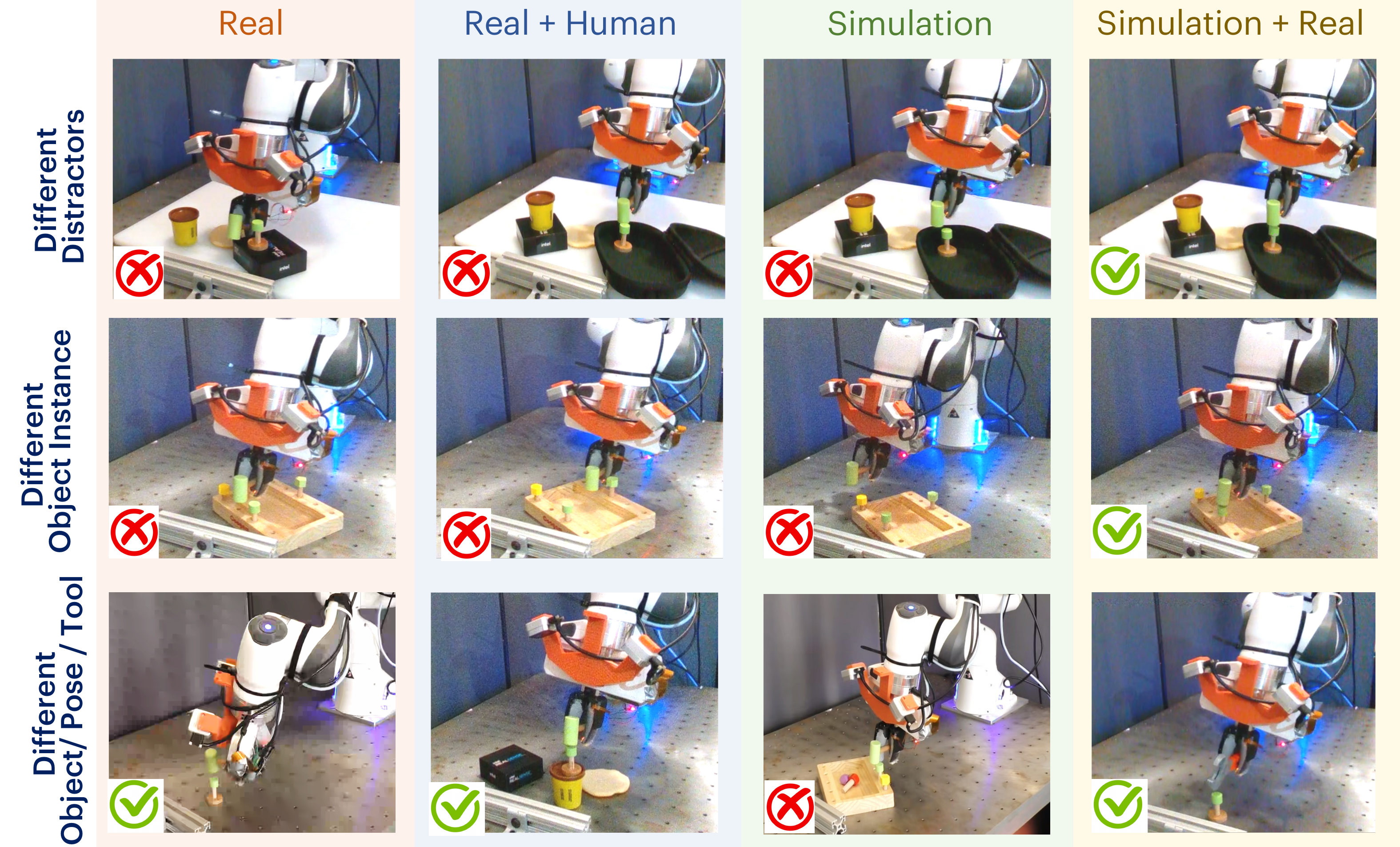}
\caption{\textbf{Domain Composition Comparison.} By composing policies trained in simulation, human, and real data, {our approach can generalize across multiple distractors (first row), with varying object and tool poses (second row), and across new object and tool instances (third row).  }}
\label{fig:scene_gen}
\vspace{-10pt}
\end{figure}

\subsection{Domain-level Compositions}

\label{sec:eval_domain_comp}
In Figure \ref{fig:sim_res} (a), we trained separate policy models for the simulation dataset $\theta_{\text{sim}}$, human dataset $\theta_{\text{human}}$, and robot dataset $\theta_{\text{robot}}$ using point cloud inputs and we evaluated domain-level composition in the simulation settings, i.e. real-to-sim evaluation when using $\theta_{\text{human}}$ or $\theta_{\text{robot}}$. To construct a policy that can work well across domains, we use domain-level composition to fuse them and evaluate them in the simulation setting. Note that there is no train/test domain gap for $\theta_{\text{sim}}$, which therefore performed well and can achieve 0.92 success rates. For example, in domains such as human data, we observe that composing with a more performant policy $\theta_{\text{sim}}$ with (e.g. $\gamma_{\mathcal{D}}=0.1$) improves the performance drastically ($\gamma_{\mathcal{D}}=0$) in simulation.  This domain level composition maps to a combined diffusion step from both policies $\epsilon^t=\epsilon_{\theta_{\text{human}}}+\gamma_{\mathcal{D}}\epsilon_{\theta_{\text{sim}}}$ in each step of the diffusion process.

\section{Real-world Experiments}
\label{sec:real_experiments}

In this section, we experiment with \METHOD{} in real-world experiments on tool-use tasks to demonstrate how data from different domains and tasks can be composed to improve generalization performance. The experiments are conducted in the same setting for each test episode across all methods. 

\subsection{Scene-level Generalization}
In this section, we experiment with composing the trained policies conditioned on observations $\mathcal{O}_{\text{realworld}}$ as RGB images $\mathcal{M}_{\text{RGB}}$ and tactile images  $\mathcal{M}_{\text{tactile}}$ in the real world, policies conditioned on observation $\mathcal{O}_{\text{human}}$ with point clouds $\mathcal{M}_{\text{pointcloud}}$ for human datasets, and policies conditioned on observation  $\mathcal{O}_{\text{sim}}$ with point clouds $\mathcal{M}_{\text{pointcloud}}$ point clouds in simulation datasets, and in real-robot datasets. We focus on the reaching task which aims to reach a pin with a hammer.

In Figure \ref{fig:scene_gen} and Figure \ref{fig:scene_variation}, we consider four common generalization axes {across the scene level in robotics in this experiment: varying object poses, varying robot initial pose, varying tool poses, adding distractor objects, and replacing the object with novel instances from the same classes. }
We note that the performance of simulation-trained point-cloud policies is inherently dependent on the image segmentation performance and point cloud quality.
In Table \ref{tab:scene_gen}, we observe that the simulation policies are robust yet additional composition with real-world policy can help when one fails under certain circumstances. Interestingly, we observe that while human data trained policies and real-robot-trained policies are not performant under different scene challenges, their compositions can outperform each individual constituent.

\begin{table}
\centering
\begin{minipage}[t]{\linewidth}
\setlength{\tabcolsep}{3pt}
\resizebox{\linewidth}{!}
{\begin{tabular}{l|ccccc}
{\bf Train/Test}   &
  {\bf Spatula} & {\bf Knife} &   {\bf Hammer}  & {\bf Wrench}  & {\bf Avg}  \\ 
  \midrule
Single-Task & 8/10    &  8/10  & 5/10    &   5/10 &  {$65\%\pm  17\%$} \\
MT Unconditioned & 6/10   &   5/10  &  5/10  &  4/10  &  {$50\%\pm 8\%$} \\
MT Conditioned & 8/10   &   5/10  &  6/10  &  2/10   & {$53\%\pm  25\%$} \\ 
MT Composition ($\alpha=0.1$) & 7/10    &   4/10  &   7/10  &  4/10 &  {$55\%\pm  17\%$} \\
MT Composition ($\alpha=2$) & 8/10    &   4/10  &   7/10  &  5/10 & {$60\%\pm 18\%$} \\
\bottomrule
\end{tabular}}

\end{minipage}
\caption{\textbf{Policy Performance on Different Tool-use Tasks.} We compare among different ways to handle multitask (MT) diffusion policy training, and find that task composition overall leads to improved performance across tool-use tasks. }
\label{tab:task_gen}
\vspace{-10pt}
\end{table}

\begin{figure}[t]
\centering 

\includegraphics[width=0.9\linewidth]{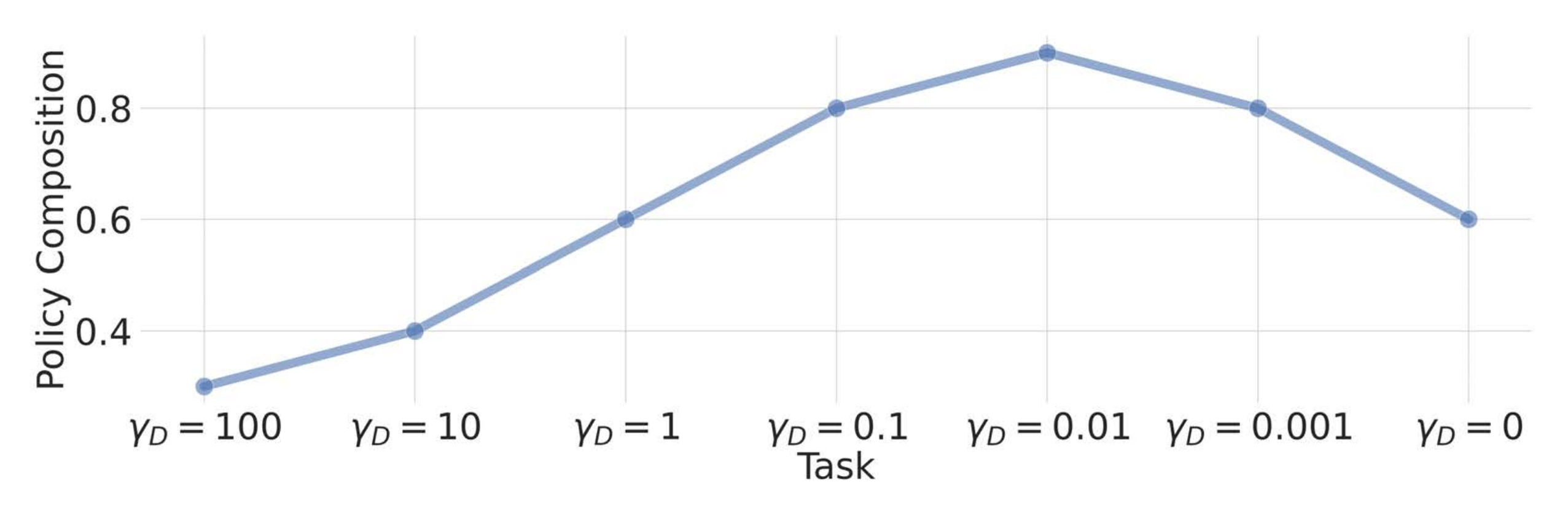}
\vspace{-12pt}
\caption{\textbf{Ablation Effect of Domain Composition Scale.} We illustrate the effect of performance on the hammering task as we increase the composition coefficient between policies.}
\label{fig:ablation_ratio}
\vspace{-13pt}
\end{figure}
\subsection{Task-level Generalization}

 We evaluate robot policy performance on four different tool-use tasks in the real world. {We call this multitask learning setting task-level generalization.} In this experiment, all policies are trained with real-world RGB and tactile data with end effector poses as input.  The multitask-conditioned policies use language features as task-specific features to concatenate with the observation features. The task-composition policies use language-conditioned classifier-free training. In Table \ref{tab:task_gen}, we found that multitask policies can perform on par with task-specific policies conditioned on $\mathcal{T}_{\text{spatula}}$ and $\mathcal{T}_{\text{hammer}}$ for example. They both showed stability in fine actions in dexterous tasks and we expect scaling up to more tasks will further improve the policy representation. We also found that language-conditioned classifier-free training performs better than naively concatenating the features when training multitask tool-use policies. We found that the composition hyperparameter needs to stay within a range to be effective and stable. Surprisingly, the trained policies exhibit robust retrying behaviors toward disturbances and robustness towards changing lighting conditions and shadows, as shown in Figure \ref{fig:realworld_robust}.

\subsection{Ablation Study}
In this section, an ablation study is performed with several components of the proposed framework. We first compare with a baseline that naively pools all data from heterogeneous domains together for learning. The policy is trained on both simulation point cloud datasets and real-world image datasets for the hammering task, by mapping both modalities into a shared latent space for the policy. 
As shown in Table \ref{tab:ablation}, this naive approach performs poorly despite having larger model size, resulting in a 75\% performance drop.
We hypothesize that the constraints and the domain gaps between the two modalities cause challenges in leveraging both modalities efficiently without a vast amount of data. In addition, we find the tactile information to be important in certain tasks that require larger gripping forces and extrinsic contact interactions such as wrench using. 
Similar to \cite{chi2023diffusion}, we also found rolling out action trajectory predictions  (e.g. 4 steps) to both improve motion smoothness and task success rates. Finally, we ablated on the composition scale (Table \ref{tab:ablation}) and found a suitable range for best performance beyond unconditional  ($\alpha=0$) and conditional ($\alpha=1$) models.


\begin{table}
\centering
\begin{minipage}[t]{1\linewidth}
\centering
\begin{tabular}{l|c|c|c}
{\bf Ablation}   &
  {\bf Data Pooling}   & {\bf No Tactile}   & {\bf No Rollout}  \\ 
   \midrule
Relative Success & $-75\%$    & $-38\%$  & $-24\%$      \\
\bottomrule
\end{tabular}
\end{minipage}
\caption{\textbf{Ablation of Real World Execution.} We analyze the effect of composition across modalities (data pooling), the use of tactile signals (no tactile) and the effect of predicting open-loop trajectory rollouts (no rollout) in the real world. {The numbers represent the relative scale of the ablated method compared to the default method that involves policy composition from simulation and real world, tactile feedback, and closed-loop predictions. } }
\label{tab:ablation}
\vspace{-20pt}
\end{table}

 \section{Limitations and Future work}
\label{appendix:limitation}

{
Our system consists of several limitations and weaknesses. In this paper, we have proposed a simple form of policy composition where multiple policy distributions are multiplied together. While such a composition is effective, it does not enable temporal composition across long-horizon tasks and it assumes policies are roughly aligned in action horizon, scale, and frequency. As a result, our paper only shows the efficacy of PoCo on short-horizon tool-use tasks, and we believe extending PoCo to temporal composition is a rich direction for future work.}

{In addition, the policies can overfit and perform poorly in tasks that require delicate contact control.
Some failure cases are illustrated in Appendix Figure \ref{fig:failure_case}.
We believe some of the failure cases can be potentially improved by (1) increased data quantity, especially higher data coverage of corner cases or failure recovery cases, and (2) reduced computational costs at the inference time, such that more individual models can be composed and/or the composed policy can be executed at a higher frequency. The current policy composition can only run at 5Hz when combining two policies, and will linearly increase when adding more policies. There are also limitations when transferring the policy trained on one tool to another tool, even in the same class. Previous works \cite{brandi2014generalizing,hillenbrand2012transferring,thompson2021shape} have several ideas that can be applied to address these tool-transfer constraints as well. }

{
Finally, the three levels of compositions in our work are only a first attempt in this direction: for task composition, we only consider 4 tasks in the multitask domains, and extending to many more tasks would be interesting. For behavior composition, we only consider analytic cost functions in the simulation environments, and adding more interesting cost functions in the real world will be interesting. For domain-level composition, we only consider normal policy models trained with in-house data, and extend to more diverse models trained with RT-X \cite{open_x_embodiment_rt_x_2023} and Ego4d \cite{grauman2022ego4d} for example would be interesting. }

\section{Conclusion} 
\label{sec:conclusion}
In this work, we proposed \texttt{PoCo}, a diffusion-based framework to compose robotic policies that tackle the data heterogeneity and task diversity problems in robotic tool-using settings.
Without any retraining, our framework can flexibly compose policies trained with data from different domains (tactile images, point clouds, and RGB images), for different tool-use tasks (scooping, hammering, cutting, and turning), with different behavior constraints (smoothness and workspace constraints).  In both simulation and the real world, we show that \texttt{PoCo} performed robustly on tool-using tasks and it showed high generalizability to various settings, compared to methods trained on a single domain. Future directions include temporal trajectory composition of different frequencies for long-horizon tasks, policy composition methods on large-scale datasets, and policy distillation from composed policies.

\section*{ACKNOWLEDGMENT}
\label{sec:acknowledgement}
Lirui Wang is supported in part by Amazon Greater Boston Tech Initiative and Amazon PO No. 2D-06310236 and Defense Science \& Technology Agency, DST00OECI20300823. Yilun Du is supported by NSF Graduate Fellowship. Jialiang Zhao is partially supported by Toyota Research Institute and Amazon Science Hub. The authors would also like to thank Sandra Q. Liu for her help in the fabrication of the hand used in this work. We thank MIT Supercloud for providing computing resources.  The authors would like to thank Max Simchowitz, Kaiqing Zhang, and William Shen for reviewing an earlier version of this draft.

\bibliographystyle{unsrt}
\bibliography{references}
\pagebreak

\clearpage 
\section{Appendix} 
\label{sec:appendix}


In the Appendix Section \ref{appendix:task}, we discuss the task descriptions and implementations of the tool-use tasks.  In Section \ref{appendix:impl}, we discuss the method implementations for training and neural networks. In Section \ref{appendix:additional}, we show additional experiments for comparing neural network architectures. In Section \ref{appendix:proof}, we derive the formulate for the policy composition.  Section \ref{appendix:dataset}, we discuss simulation, human demonstration, and real-world robot setups. In Section \ref{appendix:discussion} and Section \ref{appendix:limitation}, we discuss data heterogeneity in robotics and 
some limitations of the proposed method. 

\begin{figure*}[!t]
\centering 
\vspace{-20pt}
\includegraphics[width=0.9\linewidth]{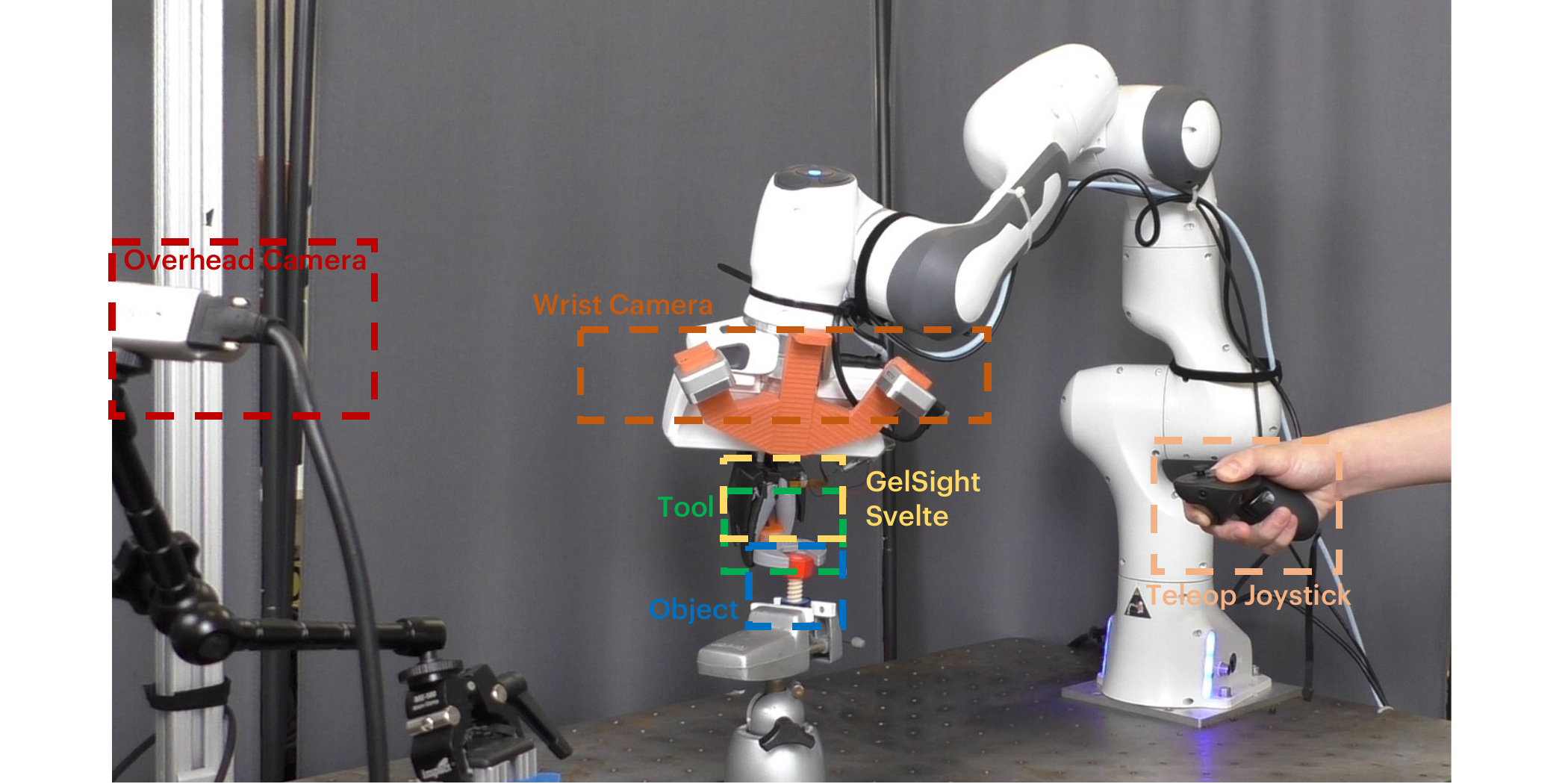}
\caption{\textbf{Real World Setup}. The robot observation are captured by a wrist camera and an overhead camera. There is a GelSight svelte hand \cite{zhao2023gelsight} mounted to the end effector grasping a tool to do certain tasks. For the teleoperation, we use an Oculus Quest Pro for tracking the joystick pose and map to the end effector velocity control.}
\label{fig:real_robot_setup}
\end{figure*}

\subsection{Task Descriptions and Implementation}
\label{appendix:task}
Although our framework can be applied to any policy setting or robotic applications, we choose 4 tasks in this work following \cite{wang2023fleet}: \texttt{use a spanner to apply wrenches to a screw}, \texttt{use a hammer to hit a pin}, \texttt{use a spatula to lift a pancake from a pan}, and \texttt{use a knife to split a play-doh}. {These tasks provide a balance of common robotic tasks that require precision, dexterity, and generality in object-object affordance. These tasks include features such as deformable object manipulation, extrinsic contacts, and forceful manipulation.} The wrench task is considered a success if the nut is turned by 15 degrees. The hammer task is considered a success if the pin is reached by the hammer. The spatula task is considered a success if the object is lifted by the spatula. The knife task is considered a success if the Play-Doh is separated in half. The hammering task is considered successful if the pin is hammered down by 2cm.  These tasks involve rich force and fine contact interactions as well as involve rigid, articulated, and deformable objects. While the hammering task is different from dynamic hammering tasks that human do, it still requires precise dexterous manipulation to execute. The real-world toy tool sets can be purchased through these links \href{https://www.amazon.com/Coogam-Construction-Montessori-Educational-Preschool/dp/B09DJY5Z26/ref=sr_1_59_sspa?keywords=tool-use+toy+set&qid=1682970991&sr=8-59-spons&psc=1&spLa=ZW5jcnlwdGVkUXVhbGlmaWVyPUFPMVFENFA0VFdQTzcmZW5jcnlwdGVkSWQ9QTA1ODQwMjFMRTZaSElUWEQzWlEmZW5jcnlwdGVkQWRJZD1BMDgwNjA0NDJaVEZaVFFTRDBTNEMmd2lkZ2V0TmFtZT1zcF9tdGYmYWN0aW9uPWNsaWNrUmVkaXJlY3QmZG9Ob3RMb2dDbGljaz10cnVl}{1},\href{https://www.amazon.com/dp/B07YJFFKHX?ref=ppx_yo2ov_dt_b_product_details&th=1}{2}.

In each of the simulation task initialization, the scene can have randomized object pose, tool-in-hand pose, and robot initial joint angles. The hammer task is considered a success if the pin is pressed down 2 cm. The wrench task is considered a success if the wrench is rotated clock-wise by more than 0.4 radians. The spatula task is considered a success if the object is lifted by over 5 cm. The knife task is considered a success if the two objects are split by over 10 cm. See Figure \ref{fig:sim_res} for successful task demonstrations. 

\subsection{Implementation Details}
\label{appendix:impl}
To implement each policy, all image inputs, including overhead camera, wrist camera, and tactile cameras are all processed with ResNet-18 \cite{he2016deep}. We apply standard color jittering and motion blurring for real-world RGB inputs. For point cloud inputs, the point clouds are transformed in the end effector frame before computing the actions. We use masked point clouds for the tool (mask 1) and the object (mask 0) and feed into a standard PointNet to compute the features. {Each encoder in each of the policies is trained separately from other datasets.} We experimented with more advanced network architectures but did not find major improvements. We apply data augmentations that maintain the object-tool relationships. For example, we change the global frames for the end effector to mimic different ways of holding the tools. We also apply noises to either the object or tool point clouds and apply corrective actions to the action labels. We also add point-wise noises, random cropping, and random dropping to the observed $512$ tool and $512$ object point clouds from the depth images and masks. 
Different from the original diffusion policy implementation \cite{chi2023diffusion}, we choose to use a fixed normalization for the action bounds rather than dataset statistics and use end-effector velocity control with six degrees of freedom parametrized in xyz and roll, yaw, pitch. {The main reason is that each of the individual policy needs to ensure their output lie in the same space so that we can compose. The original absolute coordinate frame with action bounds from dataset statistics might not have satisfy this constraint.. } We use the standard Euler angle as the rotation representation. For the joint training baselines, we map the RGB inputs and point cloud inputs to the same feature space with dimension 512 and use a policy head to regress the actions with 1-dimensional UNet \cite{janner2022planning}. The experiments use point clouds as inputs for inference time, as the RGB inputs behave quite poorly. We train all policies in simulation and the real world with 80000 steps and learning rate 0.0001 with batch size 512 and 64 respectively for around 16 hours on NVIDIA V-100.

For real-world composition, we use a fixed value of $\gamma=0.03$. We use a T5 encoder \cite{raffel2020exploring} to encode the language features and then map to dimension $256$ with a single linear layer. The drop ratio and conditional scale for classifier-free training in task composition are 0.1 and 0.01 respectively. We significantly reduce the size of the network by shrinking the embedding dimensions. We use DDPM scheduler with  100 diffusion steps in training time and DDIM scheduler with 16 steps in test time. We use the ``squaredcos\_cap\_v2'' schedule with `beta\_end=0.02`. {For tuning the composition hyperparameter, we usually start tuning by running the model with a fixed lambda weight, for instance for the classifier free inference, and then we tuned it up/down until the policies became unstable. We also monitored the numeric scale (e.g. mean absolute value of the gradient to the action trajectory) during the composition process.}

\subsection{Additional Experiments}
\label{appendix:additional}
In this section, we compare with a more standard feedforward policy parametrization. In Figure \ref{fig:add_exp} (a), we show that diffusion models can outperform MLP-based policy when testing on both in-domain tools and generalization to new tools. In Figure \ref{fig:add_exp} (b), we also showed additional comparisons on Meta-world \cite{yu2020meta}, a widely used multitask benchmark. In this case, we follow the setup in \cite{wang2023fleet} with 1000 data points for each task in MT-10 (10 tasks) with the same diffusion and policy setup as in the main Fleet-Tools experiments. We use states as policy inputs. Importantly we use the one-hot task vector in the multitask policy training.  Figure \ref{fig:scene_variation} shows the different scene generalizations in the conducted experiments.
\begin{figure*}
    \centering
\includegraphics[width=\linewidth]{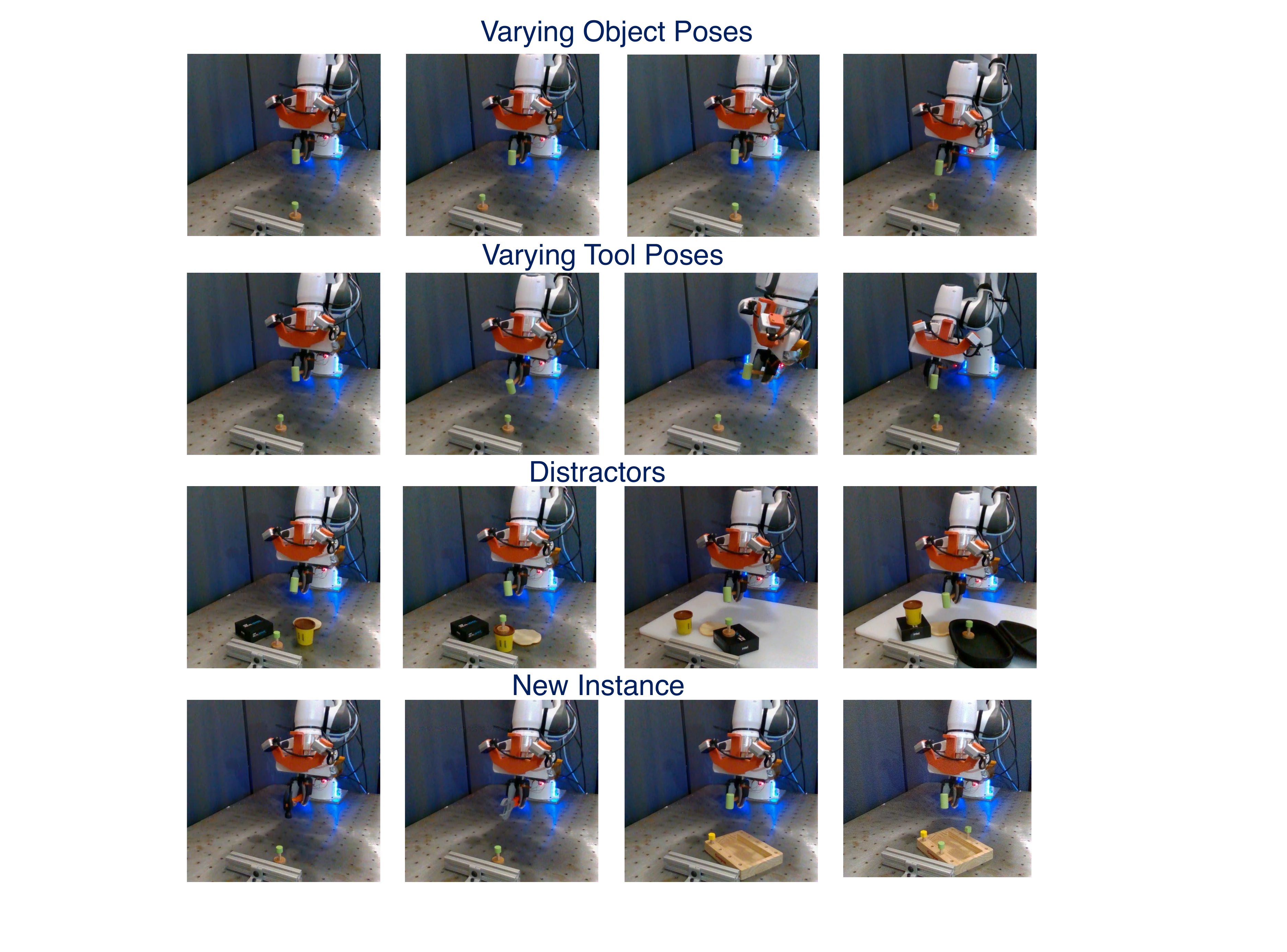}
    \caption{{A list of scene variations considered in the scene-level generalization experiments.}}
    \label{fig:scene_variation}
\end{figure*}

\subsection{Derivation and Discussion}
\label{appendix:proof}
In this section, we give a simple derivation for how the product rule of probability can be used to show Eq. \ref{eq:product}, assuming (1) mutual independence of task $\mathcal{T}$ and cost $c$ and (2) conditional independence given the trajectory. Assume that policies trained with different modalities or under different domains, or for different constraints and tasks are independent. The tasks and costs are often conditionally independent given a demonstration trajectory. Taking two separate independent sets of cost functions and tasks $A=(c_1,T_1),B=(c_2,T_2)$ for example, we want to show $p(\tau|A)p(\tau|B)=p(\tau|A,B)p(\tau)$, which shows that $p(\tau|A)p(\tau|B) \propto p(\tau|A,B)$ assuming $p(\tau)$ is uniform. To see this, note that 
\begin{align*}
p(\tau|A)p(\tau|B)&=\frac{p(\tau,A)}{p(A)}\frac{p(\tau,B)}{p(B)} \\
   &=\frac{p(A|\tau)p(\tau)}{p(A)}\frac{p(B|\tau)p(\tau)}{p(B)}  \\
   &=\frac{p(A,B|\tau)p(\tau)}{p(A)}\frac{p(\tau)}{p(B)} \\
   &=\frac{p(\tau,A,B)p(\tau)}{p(A,B)} \\
   &=p(\tau|A,B)p(\tau) \\
   &\propto p(\tau|A,B)
\end{align*}

Ideally, each cost function and task do not conflict with each other. In the case that these objectives are not independent (or orthogonal under certain structures), but where there are samples that are high likelihood under both objectives, optimizing in the product of the energy landscape will still find such plausible samples (as these samples will have the highest likelihood) in practice. 

\begin{figure}[!t]
\centering 
\includegraphics[width=\linewidth]{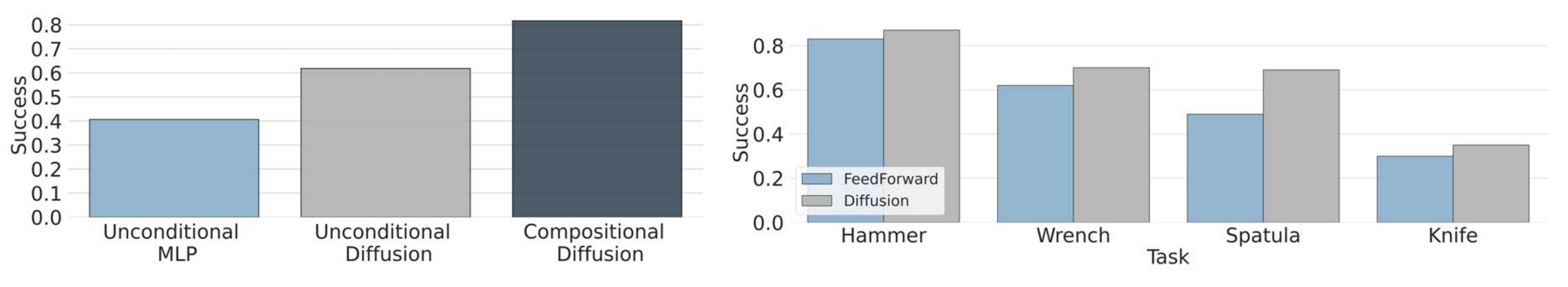}
\caption{\textbf{(a) Policy Architecture Ablation.}  Diffusion outperforms MLP policies in single-task settings in simulation. \textbf{(b) Ablation Study on Meta-world.}  We conduct additional multitask policy training experiments on the Meta-world benchmark \cite{yu2020meta}. We observed that classifier-free training can improve the performance in our setups. }
\label{fig:add_exp}
\vspace{-5pt}
\end{figure}

\begin{figure*}[!t]
\centering 
\vspace{-25pt}
\includegraphics[width=0.8\linewidth]{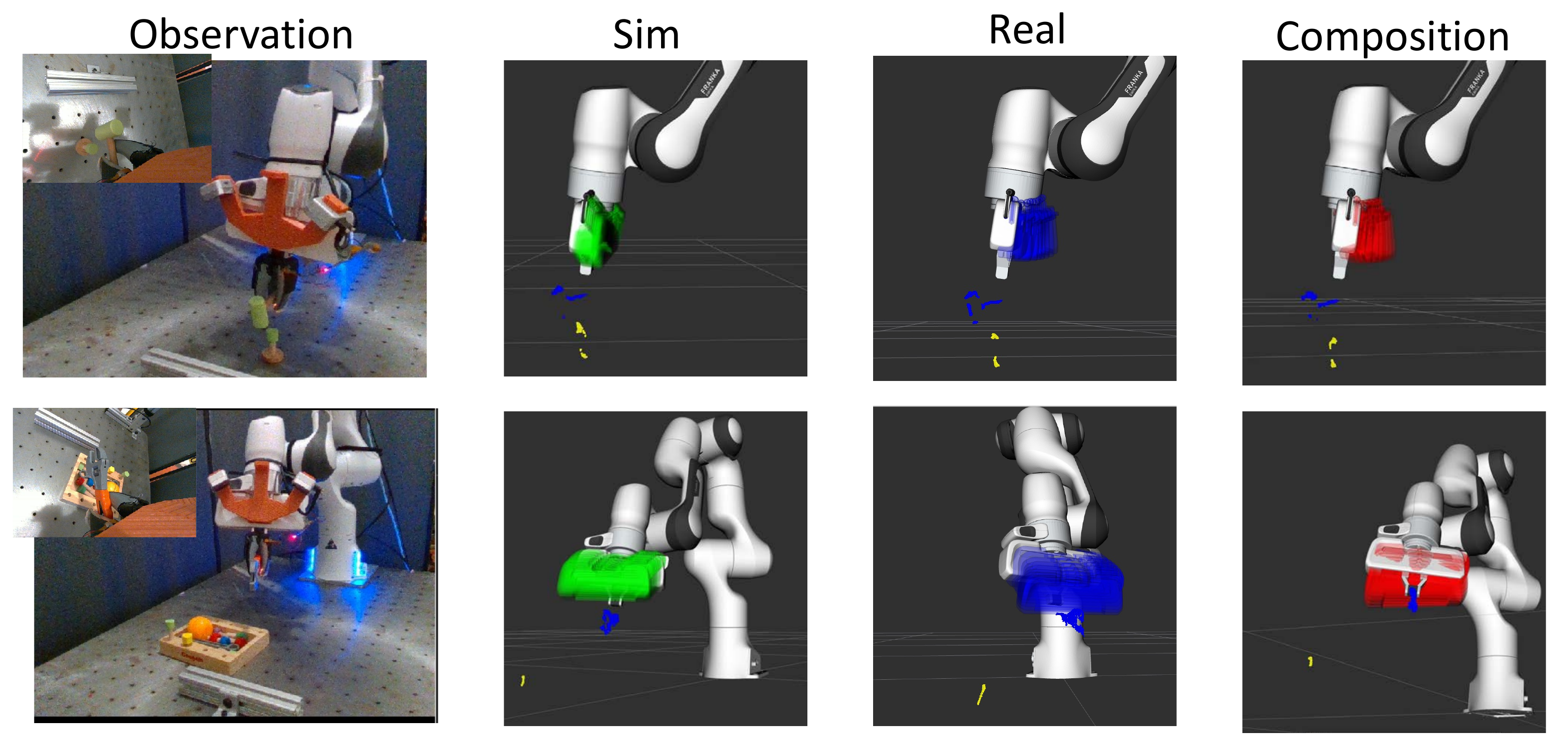}
\caption{\textbf{Trajectory Composition Snapshots.} The left frame shows the overhead view and the wrist view and the rest three frames show different trajectories predicted. The top row shows that when the objects are very close by, which causes partial observations in depth sensing and segmentation, the simulation policy itself can generate the wrong action trajectories (going forward instead of backward). On the other hand, when the scene drastically changes to a cluttered scene and the tool is changed, the RGB policy can also generate wrong motions. In both cases, the policy composition improves the performance of each constituent policy. }
\label{fig:vis_single_frame}
\end{figure*}


\subsection{Dataset Collection and Robot Setup}
\label{appendix:dataset}

\subsubsection{Human Dataset}
Figure \ref{fig:data_processing} shows our pipeline to process human demonstration data into point cloud action trajectories. We use standard off-the-shelf libraries such as segment-anything \cite{kirillov2023segment} and  X-Mem \cite{cheng2022xmem} for image segmentation and video tracking. We build a simple customized interface to label points on the image in batch and then use the point prompt in segment-anything \cite{kirillov2023segment}  to extract a single-frame mask. {The Segment Anything Model (SAM) pipeline offers various forms of prompts, one of which is a point prompt, which allows us to use a few mouse clicks to initiate a mask prediction for a certain object or tool from the model. After that, the model would output masks in three levels of granularity, and we will pick the middle one as the output mask for this particular instance.}
 The mask is then propagated into XMem \cite{cheng2022xmem}  for video mask tracking. The mask together with the depth image are used to compute the masked point cloud. The masked point clouds in two frames can be used to compute a relative action using the iterative closest point (ICP in the open3d library). We visualize masked videos and segmented point clouds with action trajectories. Note that no camera calibration is needed and each camera provides an independent data point. The human data collection phase can collect up to 200 trajectories for 20 minutes. One limitation is the quality of the depth point cloud and the distribution shifts from the overhead camera that records the human demonstrations and the wrist camera that is used in robot evaluation time. One way to resolve this is to use camera mounts on human wrists when collecting the data and computing the relative poses via SLAM.

\subsubsection{Real Robot Dataset}
The robot setup has two active cameras (one overhead Intel D-435 and one wrist camera D-405). {The overhead camera is put in front of the robot to have a global view of the robot and the scene and to avoid single-view ambiguity in the policy execution process. The wrist camera view is useful to provide more generalization [1] for the policies as shown in previous works. 
} The robot uses a GelSight Svelte \cite{zhao2023gelsight} fingers to hold a tool (wrench in this case) to activate an object (nut on the bolt). {For the Gelsight Svelte, we note that it’s a 3-finger hand that provides a firm grasp of many tool types, and also tactile information is useful for many aspects including force/tactile feedback in the tasks, locating tool-in-hand poses, and robust to environmental changes such as lighting.} The RGB policy frequency is 10 and the point cloud perception pipeline frequency is around 5. The end effector and joint position state frequencies are 1000Hz. The tactile image is 30Hz.  The tool and object poses are slightly varied for each episode. The real-world point clouds are sampled using farthest point sampling and are accumulated across 5 frames between every reset. After the robot commands the end effector desired pose, the low-level controller sends target joint angles asynchronously and generates a joint-space position trajectory. The end-effector frame action space and the joint velocity are limited to 3 cm per step. We collect between 50 to 100 trajectory demonstrations for each task. 

For real-world data collection (Figure \ref{fig:real_robot_setup}), we use Oculus Quest Pro to teleop the robot to collect demonstrations. We use standard world-frame velocity control and use extrinsic rotations. The teleop interface also has a switch for re-adjusting the demonstrator's pose and tracking frame reset and finger control functions. We use \cite{drake} to solve for Inverse Kinematics of the tracking frame and compute joint-space trajectories asynchronously. The robot controller takes into account the table heights when solving for the inverse kinematics. The initialization for each trajectory by default only differs by human resetting errors, while for each generalization axis, we control the experiment variable and change corresponding factors, such as adding a distractor or moving the tool.


\begin{figure}[!t]
\centering 
\vspace{-20pt}
\includegraphics[width=\linewidth]{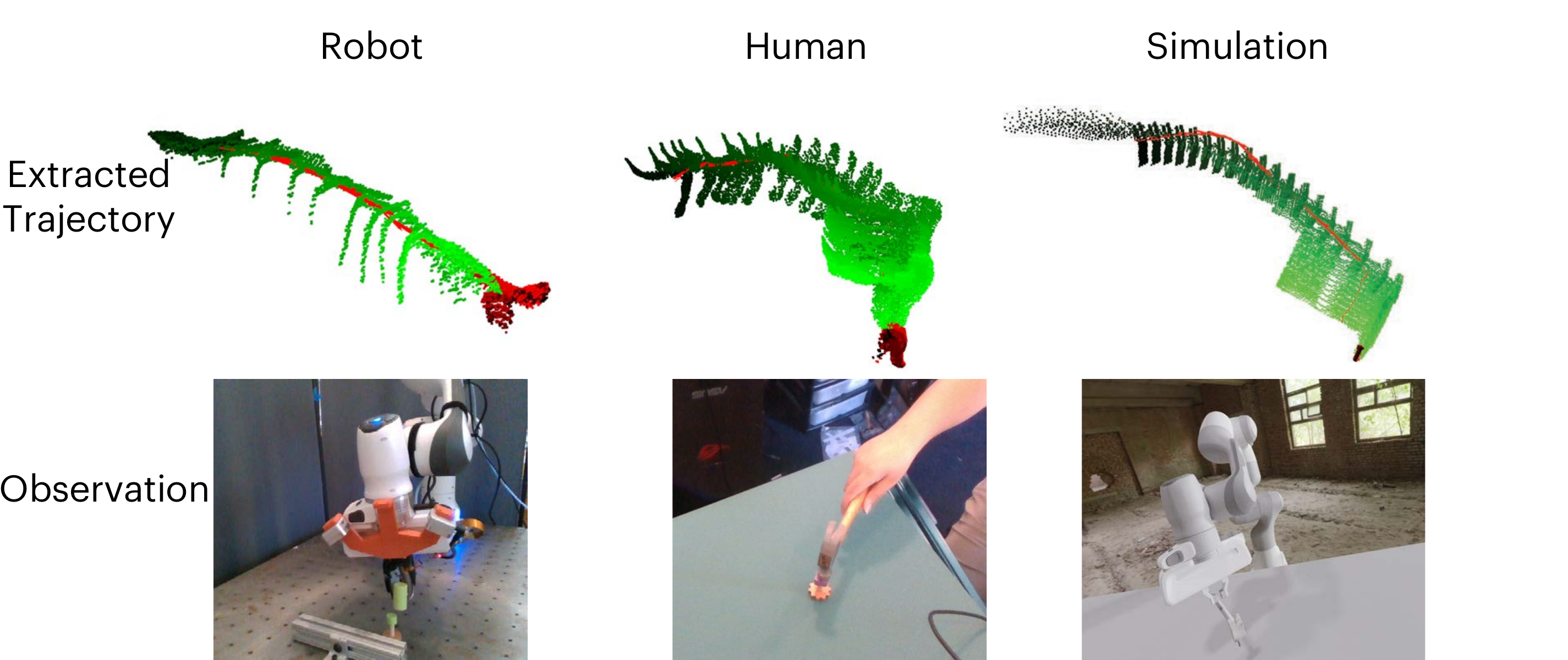}
\caption{Comparison of extracted point cloud trajectory (with actions) across different domains. We note that the human data requires no calibration and can be conducted in the wild. The datasets can be collected and processed efficiently.}
\label{fig:pointcloud_compariosn}
\end{figure}

\subsubsection{Simulation Dataset}
The simulation datasets follow Fleet-Tools \cite{wang2023fleet}  where the expert demonstrations are generated with keypoint trajectory optimizations. More specifically, we label $3$ keypoints on the tool and $1$ keypoint on the object and then use joint-space KPAM \cite{wang2023fleet} to find a joint-space trajectory for solving each of the tool-use tasks. In total, we collected around 50,000 simulated data points for each of the 5 tool-object pairs across 4 tool families. Each trajectory has frequencies around 6Hz and a length around 25. We apply domain randomization to object poses, tool-in-hand poses, and initial joints in each scene and save fixed scenes for testing. {The object models are found in  ShapeNet and Objectverse. The language annotations are labeled by humans and the datasets are combined together for multitask training.}

\subsection{More Discussions on Heterogeneity}
\label{appendix:discussion}
Robots and embodied agents inherently have to deal with heterogeneous inputs and outputs due to the nature of the perception-action loops across diverse environments. We discuss the heterogeneity of data domains in robotics. There are many reasons for this heterogeneity, one of which is different hardware naturally generates slightly different data. In particular, there are three main sources of data domains: simulation, real-robot, and human data. We discuss as follows:
\begin{enumerate}
    \item \textbf{Simulation Data.} Simulation data \cite{zhao2020sim}  has the promise of providing huge amounts of diverse data and yet suffers from sim-to-real gaps and the effort and challenges of building complex simulation tasks such as deformable objects. Moreover, simulation provides a useful environment for evaluation and for measuring progress. 
    \item \textbf{Human Demonstration Videos.} Real-world human demonstrations, such as \cite{grauman2022ego4d}, are the biggest existing data source available online for training robots (YouTube for instance) and are easy to acquire. This is important to achieve some surprising results from the success of computer vision and natural language processing. Yet, it suffers from embodiment gaps, especially on contact-rich motions. 
    \item \textbf{Real-world Robot Data.} On-robot data, such as \cite{open_x_embodiment_rt_x_2023},  has the least domain shifts, yet it can be prohibitive to acquire at large scales. Even with on-robot data, there is a difference between other robots that share training data and the robot that does the evaluation. For instance, \cite{open_x_embodiment_rt_x_2023} has 22 embodiments and there are different hardware, sensors, and environments associated with each dataset.
\end{enumerate}

Next, we provide a short discussion on some of the heterogeneity in data modalities for robots that interact with the physical world.
\begin{enumerate}
    \item \textbf{Image/Video Data.} RGB Image data is ubiquitous and has the redundancy necessary for representation learning to extract useful structures. It also has the regular data formats and engineering stacks to work with. In the context of robotics, the main issue is that policies learned from RGB data only can be brittle to irrelevant details such as lighting and backgrounds.
    \item \textbf{Depth/Point Cloud Data.} Depth and the associated point cloud are lightweight to work with. It allows a simple fusion of history and multiple views and simpler sim-to-real transfer. The problem is that the point cloud will require segmentation and the depth camera can have issues with thin and transparent objects.
    \item \textbf{Tactile/Haptic Data.} Haptic and tactile data provide useful intrinsic information about the contacts and the fine details. Human uses this information all the time during manipulation. Hardware durability and reproducibility are common issues.
    \item \textbf{State Data.} State data can contain robot proprioceptive information such as end effector poses, joint angles, velocities, and external torques, which are useful information that summarizes the state of the robots. It can also contain estimated object poses and class information as an abstraction of the world. The problem is that the perception errors will propagate to robot policies.
    \item \textbf{Language/Goal Data.} Language and other abstracted goal data (such as goal frame) provide useful intent for each task that the robot is commanded to achieve. Yet, it usually misses the low-level details of how to achieve these goals. 
\end{enumerate}
 Due to the absence of an internet-equivalent and homogeneous dataset for language models in robotics, we believe that the data heterogeneity needs to be considered carefully to use all of such data. Taking the common data modalities RGB and depth, for example, a naive data pooling approach would require paired data. Given two separate depth and color datasets, it's likely to bottleneck the overall dataset size by the dataset with the smaller data amounts. In contrast, our method aims to combine the best of all worlds. There are additional challenges such as data dependence within an episode, data distribution shifts, and long-tail problems, as well as task-level differences. Policies trained with different domains and modalities also can perform differently due to their training distributions. For example, RGB policies can generalize well for a single dexterous task in a fixed scene given enough data but can fail to generalize across details such as lighting. Depth / point-cloud policies trained in simulation can generalize across partial views and randomized scenes but can be challenging in corner cases and contact-rich scenarios. Empirically, the simulation-trained policies also behave less smoothly due to the domain gaps. Tactile-based policies can be useful in contact-rich tasks but require specific data collection procedures.

\begin{figure}[!t]
\centering 
\includegraphics[width=\linewidth]{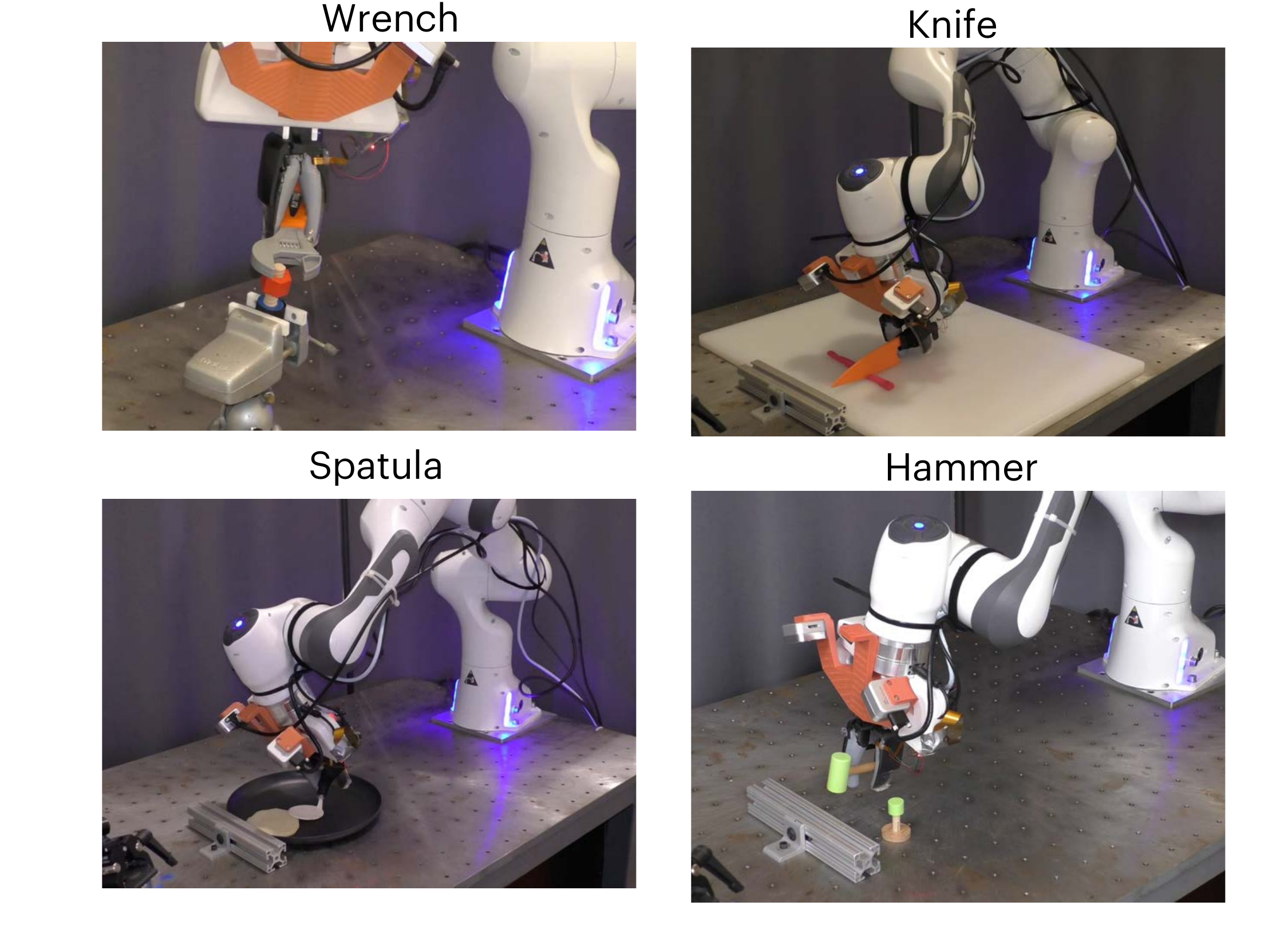}
\caption{\textbf{Failure Cases}. We have seen some failure cases in fine-grained tasks such as the wrench task and bad contact interaction such as in the spatula and knife tasks. }\vspace{-10pt}
\label{fig:failure_case}
\end{figure}


\end{document}